%% file: main.tex
\documentclass[10pt,twocolumn,letterpaper]{article}

\usepackage{cvpr}              % To produce the CAMERA-READY version
\input{preamble}

\definecolor{cvprblue}{rgb}{0.21,0.49,0.74}
\usepackage[pagebackref,breaklinks,colorlinks,citecolor=cvprblue]{hyperref}
\usepackage{multirow}  
\usepackage{caption}
\usepackage{subcaption}
\def\paperID{3946} % *** Enter the Paper ID here
\def\confName{CVPR}
\def\confYear{2024}

\title{Fully Exploiting Every Real Sample: SuperPixel Sample Gradient Model Stealing}

\author{
Yunlong Zhao$^{1}$
\quad
Xiaoheng Deng$^{1}$
\thanks{Corresponding author is Xiaoheng Deng}
\quad
Yijing Liu$^{2}$
\quad
Xinjun Pei$^{1}$
\quad
Jiazhi Xia$^{1}$
\quad
Wei Chen$^{2}$\\
$^{1}$Central South University \quad $^{2}$Zhejiang University \\
\hspace{-0.4cm}{\tt\small \{zhaoyl741, dxh, pei\_xinjun, xiajiazhi\}@csu.edu.cn,\{liuyj86, chenvis\}@zju.edu.cn}
}
\begin{document}
\maketitle
\input{sec/0_abstract}    
\input{sec/1_intro}
\input{sec/2_related}

\input{sec/3_SPSG}
\input{sec/4_Experiment}
\input{sec/5_conclusion}

\clearpage
{
    \small
    \bibliographystyle{ref}
    \bibliography{ref}
}

% WARNING: do not forget to delete the supplementary pages from your submission 
\input{sec/X_suppl}

\end{document}

%% file: preamble.tex
\usepackage[dvipsnames]{xcolor}

%% file: sec/0_abstract.tex
\begin{abstract}

Model stealing (MS) involves querying and observing the output of a machine learning model to steal its capabilities. The quality of queried data is crucial, yet obtaining a large amount of real data for MS is often challenging. Recent works have reduced reliance on real data by using generative models. However, when high-dimensional query data is required, these methods are impractical due to the high costs of querying and the risk of model collapse. In this work, we propose using sample gradients (SG) to enhance the utility of each real sample, as SG provides crucial guidance on the decision boundaries of the victim model. However, utilizing SG in the model stealing scenario faces two challenges: 1. Pixel-level gradient estimation requires extensive query volume and is susceptible to defenses. 2. The estimation of sample gradients has a significant variance. This paper proposes Superpixel Sample Gradient stealing (SPSG) for model stealing under the constraint of limited real samples. With the basic idea of imitating the victim model's low-variance patch-level gradients instead of pixel-level gradients, SPSG achieves efficient sample gradient estimation through two steps. First, we perform patch-wise perturbations on query images to estimate the average gradient in different regions of the image. Then, we filter the gradients through a threshold strategy to reduce variance. Exhaustive experiments demonstrate that, with the same number of real samples, SPSG achieves accuracy, agreements, and adversarial success rate significantly surpassing the current state-of-the-art MS methods. Codes are available at \hyperlink{https://github.com/zyl123456aB/SPSG_attack}{https://github.com/zyl123456aB/SPSG\_attack}.
\end{abstract}

%% file: sec/1_intro.tex
\section{Introduction}
\label{sec:intro}

Machine Learning as a Service (MLaaS), enhances efficiency in both work and daily life \cite{MLaaS1,han2022inet,zhizhong2024point,peng2023visual}. These invaluable MLaaS models have become targets for malicious users to steal. Model Stealing (MS) \cite{activethief,blackdissector,blackripper,delvingMS,DFME,DFMS,DS,EDFBA,IDEAL,knockoff,ZSDB3KD,CloudLeak} involves constructing a proxy model similar to the victim model by acquiring query results of input samples. Beyond the direct utilization of the proxy model, malicious users can also generate a series of attacks based on the proxy model and transfer them to the victim model, including membership inference attacks \cite{membership}, adversarial attacks \cite{adversarialP}, and model inversion attacks \cite{modelInversion}, etc.

The basic paradigm of MS is to construct a sample attack set and train a proxy model through the samples in the attack set and the corresponding query results of the victim model. Data-free MS is based on generative networks and uses noise to synthesize artificial images for the training of the proxy model. Although data-free MS claims that real samples are not needed, in practical applications that require high-dimension samples, data-free MS still has the inevitable real sample demand. As shown in Table \ref{functional}, on the one hand, high-dimensional inputs would significantly increase the query volume for data-free MS. For color images of 224x224 pixels, the query volume for data-free MS could reach tens of millions. On the other hand, due to the inherent risk of model collapse in GANs \cite{GAN}, training proxy models with data-free MS is prone to failure. By giving data-free MS a small amount (10k to 20k) of real samples related to the domain of the victim model as image priors, data-free MS can reduce the query cost and model collapse risk and improve the stealing effect. However, acquiring high-quality real samples that meet specific MLaaS requirements is challenging. Different MLaaS have varying requirements for input images, and blurred input images can distort MLaaS query results. Privacy and copyright protections further complicate the acquisition of high-quality real samples. Additionally, even with a plethora of real samples, the marginal benefit of each real sample for improving the proxy model diminishes as the cardinality of real samples increases. Therefore, expensive real samples should be fully utilized.
\begin{table*}[htbp]
    \centering
    \caption{Results of data-free MS without using real samples and with using 10k domain-relevant real samples. Results are queries, real samples, the failure times, and test accuracy (in \%), of each method with querying probability. The failure times are determined by the number of model collapses observed over 10 training runs. we report the other average result computed over 10 runs. The training strategies and experimental configurations used are described in the experimental section \ref{exp}. (1k=1000)}
    \resizebox{0.9\textwidth}{!}{
    \begin{tabular}{|l|ccc|ccc|ccc|ccc|}
    \hline
    \multirow{2}{*}{Data-free MS(probability)} 
    & \multicolumn{3}{c|}{CUBS200(0)}          & \multicolumn{3}{c|}{Indoor67(0)} &\multicolumn{3}{c|}{CUBS200(10k)}          & \multicolumn{3}{c|}{Indoor67(10k)}\\
    \cline{2-13}
     & queries  & failures &accuracy & queries  & failures &accuracy& queries  & failures &accuracy& queries  & failures &accuracy\\
    \hline
    ZSDB3KD \cite{ZSDB3KD}           & 5183k  & 3    & 48.72 & 5771k  & 4 & 51.77   & 1202k\textcolor{green}{$\downarrow$}   & 2\textcolor{green}{$\downarrow$}    & 51.47\textcolor{green}{$\uparrow$} & 1112k\textcolor{green}{$\downarrow$} & 2\textcolor{green}{$\downarrow$} & 59.41\textcolor{green}{$\uparrow$}   \\
    DFMS \cite{DFMS}   & 4987k      & 2    & 51.28 & 4771k &2 &55.21 & 1021k\textcolor{green}{$\downarrow$}    & 1\textcolor{green}{$\downarrow$}    & 55.21\textcolor{green}{$\uparrow$} & 1009k\textcolor{green}{$\downarrow$}  &0\textcolor{green}{$\downarrow$} &61.11\textcolor{green}{$\uparrow$}   \\
    EDFBA \cite{EDFBA}   & 1623k      & 2    &51.46  & 1792k  &1 &56.27   & 459k\textcolor{green}{$\downarrow$}     & 0\textcolor{green}{$\downarrow$}    &55.61\textcolor{green}{$\uparrow$} & 349k\textcolor{green}{$\downarrow$}  &0\textcolor{green}{$\downarrow$} &61.17\textcolor{green}{$\uparrow$}\\
    DS \cite{DS}    & 4310k     & 3    & 50.26 & 4291k  &2 &54.21  & 1077k\textcolor{green}{$\downarrow$}    & 1\textcolor{green}{$\downarrow$}    & 54.97\textcolor{green}{$\uparrow$} & 1021k\textcolor{green}{$\downarrow$}  &1\textcolor{green}{$\downarrow$} &60.27\textcolor{green}{$\uparrow$}  \\
    DFME \cite{DFME}   & 4102k   & 2   & 48.82 & 4671k  &2 &52.78 & 1489k\textcolor{green}{$\downarrow$}  & 1\textcolor{green}{$\downarrow$}   & 52.31\textcolor{green}{$\uparrow$} & 1339k\textcolor{green}{$\downarrow$} &1\textcolor{green}{$\downarrow$} &58.71\textcolor{green}{$\uparrow$}  \\

    \hline
    \end{tabular}}
    \label{functional}
\end{table*}

To make full use of each real sample, our idea is to obtain more information about the model from real samples. However, for black-box models, it is challenging to obtain other types of information to train proxy models.
\begin{figure}[h]
    \centering
    \includegraphics[width=\columnwidth]{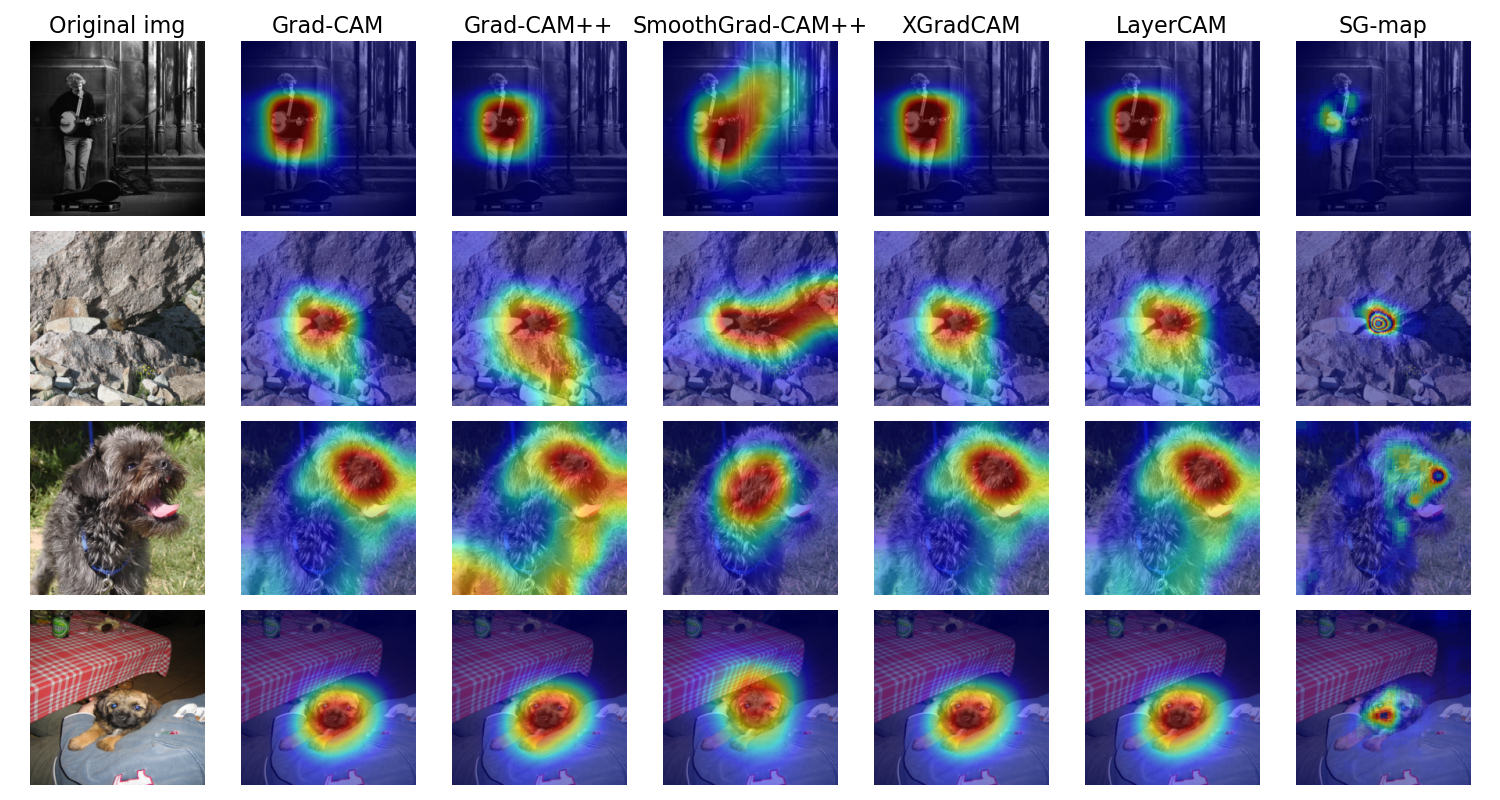}
    \caption{The columns from left to right are grad-CAM \cite{gradCAM}, grad-CAM++ \cite{gradcam++}, Smooth-gradCAM \cite{smoothgradcam++}, X-gradCAM \cite{xcam}, layer-CAM \cite{layercam}, and SG-map. The neural network is ResNet34 pre-trained on ILSVRC-2012.}
    \label{SG-map}
\end{figure}
For example, all feature maps from the intermediate layers of the black-box model \cite{kdf1,kdf2,kdf3,kdf4} are unknowable. Our focus shifts to the input of the model, specifically, the Sample Gradient (SG) backpropagated from the model’s output layer to the input sample. SG is used to assist in generating the interpretability heat map of the model in many interpretability works. Actually, SG itself contains a lot of model interpretability decision information. As shown in Figure \ref{SG-map}, we compare the heat map generated by average pooling SG with other CAM methods, and we can see that SG fully reflects the model’s decision. However, the obstacles for imitating sample gradients in MS include:
(1) The most primitive method to obtain sample gradients involves perturbing each pixel individually and acquiring query results for all perturbed images, a process with enormous query costs. Specially, obtaining sample gradients for a single image requires over a hundred thousand queries. Furthermore, inputting perturbed images at the pixel level into the victim model is akin to inputting adversarial images, which could be easily detected and thwarted by defense mechanisms like Prada \cite{prada}. (2) Sample gradients have significant variance due to certain specific or redundant neurons backpropagating. Then, we introduce a novel SuperPixel \cite{quickshift,felzenszwalb,slic} Sample Gradient Model stealing  (SPSG) to solve the issues. SPSG comprises two modules: superpixel gradient querying (SPGQ) and sample gradient purification (SGP). For issue (1), SPGQ module first segments the image into multiple superpixels based on a segmentation algorithm. Then, it applies perturbations to these superpixels and queries the output to obtain the sample superpixel gradients. For issue (2), SGP module eliminates significant variance from the sample gradients by filtering extremum information and normalizing, ensuring the extraction of clean and useful gradient information. Then, purified superpixel gradients are associated with the pixel gradients of the proxy model to train proxy models. 

Our contributions are enumerated as follows:
\begin{itemize}
    \item We design SPSG to extract the maximum amount of available model information from each real sample. The superpixel querying module significantly reduces the query volume required to acquire the tacit knowledge in one sample gradient from $10^6$ to $10^2$ while simultaneously evading defenses like Prada \cite{prada}. Meanwhile, the gradient purification module effectively removes noise from the sample gradients. The effectiveness of the gradient purification module is further validated through ablation experiments. 
    \item Through various experiments, SPSG enables the proxy model to achieve accuracy, agreement, and attack
    success rate substantially surpassing state-of-the-art algorithms with the same number of real samples. Specifically, when stealing a resnet34 model trained on CUBS-200 using 20,000 real samples, SPSG achieves an accuracy of 61.21\% and an agreement of 67.48\%, significantly outperforming the second-best method with an accuracy of 56.39\% and agreement of 58.44\%. To match the accuracy achieved by SPSG with 10,000 real samples, the second-best method requires at least 20,000 real samples.
\end{itemize}

%% file: sec/2_related.tex
\section{Related work}
\subsection{Sample Gradient}
Sample gradients, obtained through the backpropagation of a model's final loss function, depend on the parameters and structure of the neural network. They are primarily used in adversarial training and model interpretability.

\textbf{Adversarial Training Based on Sample Gradients.} By introducing small perturbations along the direction of sample gradients, new samples capable of deceiving the model can be generated. Techniques like FGSM \cite{FGM} are sensitive to the sign of sample gradients, whereas FGM \cite{FGSM} normalizes the gradients. PGD \cite{PGD} and FreeAT \cite{freeAT} implement multiple iterations with smaller step sizes on sample gradients, keeping the perturbations within a specified range. YOPO \cite{YOPO} reduces gradient calculation costs by leveraging the network's structure, with perturbations related only to the first layer. FreeLB \cite{freeLB} accumulates gradients during training, giving perturbations a more directional tendency.

\textbf{Model Interpretability Based on Sample Gradients}. Techniques like "Saliency Map" \cite{sliencemap} create saliency maps by calculating gradients of input images to highlight the model's focus areas in image classification tasks. Guided Backpropagation \cite{Striving} helps understand the decision-making process and the features learned by each convolutional layer. GRAD-CAM \cite{gradCAM} and SmoothGrad \cite{Smoothgrad}, while not directly using sample gradients, generate heatmaps using feature map gradients to show the focused image areas. \cite{Interpretableblack} proposes a method for explaining black-box models through input perturbations, generating interpretability masks to illustrate the model's focus areas.

\subsection{Superpixel Segmentation}
Superpixels were introduced to create image over-segmentation based on similarity criteria. Algorithms like SLIC \citep{achanta2012slic} efficiently generate superpixels by clustering pixels in a five-dimensional color and image plane space. Subsequent methods \cite{felzenszwalb,quickshift} have improved superpixel segmentation.

\subsection{Model Stealing}
\textbf{Data-Free Model Stealing.}
Data-free model stealing techniques \cite{blackripper,delvingMS,DFME,DFMS,DS,EDFBA,IDEAL,ZSDB3KD}  do not require any original training data. Attackers generate synthetic queries, often through prior knowledge or assumptions about the data distribution, to probe the model and reconstruct its functionality. All data-free model stealing (MS) inevitably draws on the concept of Generative Adversarial Networks (GANs). Therefore, there is an unavoidable risk of model collapse. Given that the querying cost required for a single instance of model theft is quite high, a collapse during the extraction process would further increase the querying costs. Additionally, since the training and querying phases in data-free MS are coupled, each training of a proxy model requires a new round of queries, which also increases the query volume. 

\textbf{Data-Driven Model Stealing.}
Data-driven model stealing attacks \cite{activethief,blackdissector,knockoff}, utilize real data, allowing for all attack set samples to be queried before training the proxy model. Therefore, it is not necessary to query the victim model again with each training of a new proxy model. The real samples can be domain-irrelevant to the victim model. Although domain-relevant real samples can achieve certain improvements in the effectiveness of the theft, domain-irrelevant real samples are more commonly used. Data-driven MS can also further refine the selection of real samples based on information from the training process of the proxy model to enhance the stealing effect.

%% file: sec/3_SPSG.tex
\section{SuperPixel Sample Gradient Model Stealing}
\subsection{Overview}
\begin{figure*}[h]
    \centering
    \includegraphics[width=0.75\textwidth]{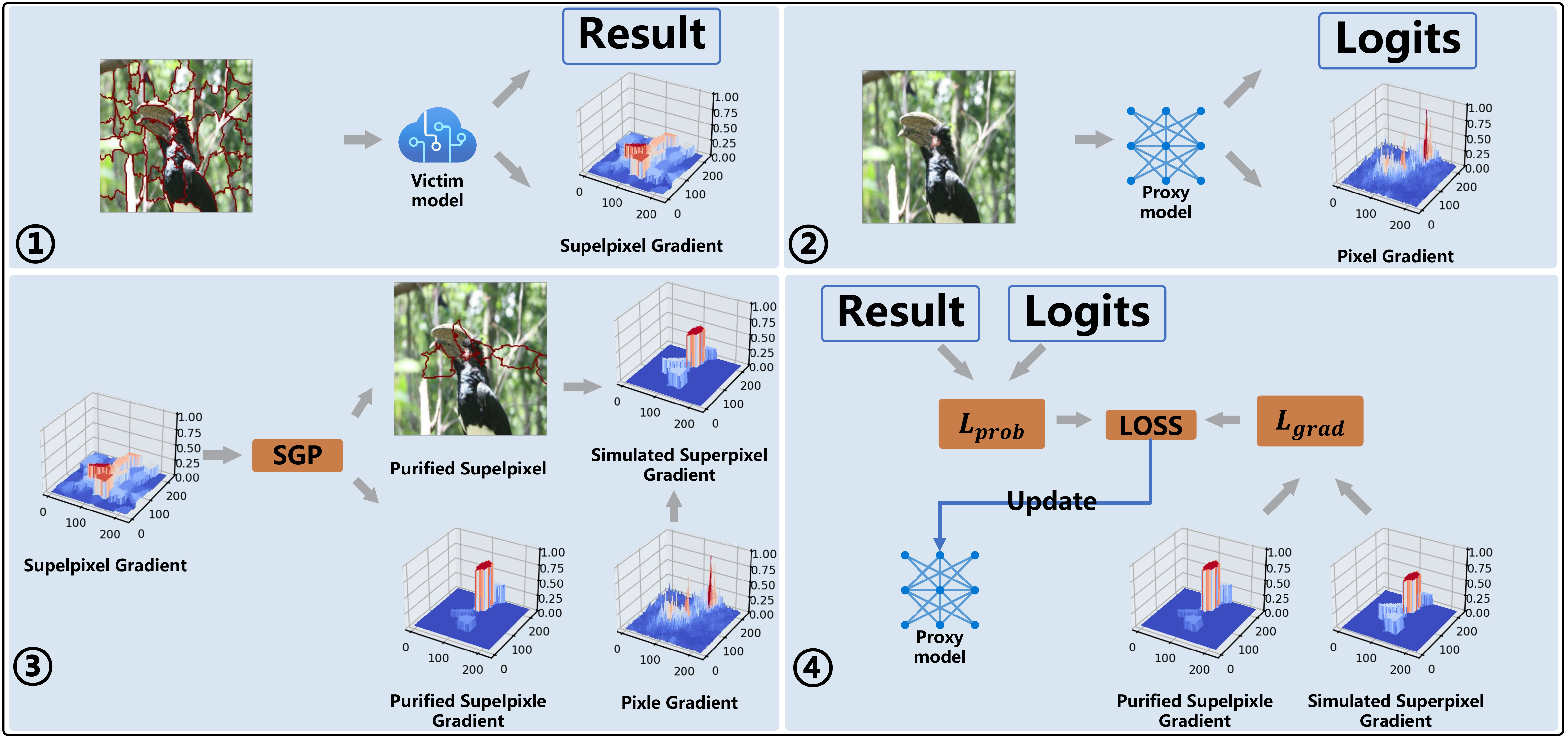}
    \caption{Four steps of SPSG. The first step is to obtain superpixel gradients and query results through SGPQ. The second step involves acquiring pixel gradients and output logits of the proxy model through backpropagation on the input sample. The third step is to obtain purified superpixel gradients and simulated superpixel gradients of the proxy model using SGP. The fourth step involves updating the proxy model based on the loss function. The gray arrow represents the direction from input to output.}
    \label{SGSP}
\end{figure*}

In the fundamental paradigm of offline Model Stealing (MS), MS begins with the construction of a query set comprising all input samples and their corresponding query results. Subsequently, this pre-assembled query set is utilized to train different proxy models. SPSG, falling under the category of offline MS, mainly encompasses two distinct modules: SuperPixel Gradient Query (SPGQ) and Sample Gradient Purification (SGP), as shown in Figure \ref{SGSP}.
SPGQ is designed for the assembly of the query set. In addition to acquiring the predictive probabilities or hard labels for each sample’s output, SPGQ is also adept at obtaining superpixel gradients for each sample at a low query cost, while simultaneously circumventing adversarial attack monitoring. On the other hand, SGP is employed for the training of the proxy model. Within this module, the superpixel sample gradients from the query set undergo a denoising process. This process ensures the retention of the extremal portions of the superpixel gradients across each channel of the image. Based on every superpixel range filtered by the victim model, the pixel gradients of the proxy model are averaged to obtain the simulated superpixel gradients. In the final stage, we introduce a novel loss function that establishes a connection between simulated superpixel gradients and their ground-truth. Through this connection, the proxy model is effectively trained, culminating in a comprehensive and robust offline MS framework.

\subsection{SuperPixel Gradient Query}
For black-box models, the finite difference method for calculating pixel gradients of input images is a primitive yet effective approach. The finite difference method estimates gradients by applying a small perturbation to the input sample and observing the resultant output changes. Finite difference includes forward difference, central difference, and backward difference. In this paper, we default to using the forward difference. Specifically, for an input sample \( x \) and a small perturbation \( \varepsilon \), an approximation of the gradient for each pixel \( i \) in channel \(c = 1 \vee 2 \vee 3\) can be calculated using the following formula: \begin{equation}
   q_i^c =\frac{\partial f}{\partial x} \approx \frac{f(x + \varepsilon \cdot e_i^c) - f(x )}{\varepsilon}
  \label{eq:finite difference}
\end{equation} 
Here, \( \varepsilon \) must be sufficiently small to capture the function's variations at \( x \), yet not too small to avoid numerical precision issues. In this paper, the value of \( \varepsilon \) is set to \( 1e-5 \).  \(e_i\) is a standard basis vector with only the i-th pixel in channel c as 1. However, pixel-level forward differences require one perturbed query per input dimension, rendering this method computationally expensive in high-dimensional input spaces. In addition, MLaaS can track and store a series of pixel-level perturbed images of the input. Under normal circumstances, the pairwise distance between queried images with the same label follows a Gaussian distribution. However, with the addition of tiny perturbations, the pairwise distance no longer follows a Gaussian distribution and tends to an extreme distribution. Prada \cite{prada} can detect such changes in distribution and outputs a noisy prediction result upon detection of change.

Therefore, we propose SuperPixel Gradient Query (SPGQ), whose core idea is to extend the pixel-level forward difference to the superpixel level. Superpixels are a concept in image processing and computer vision, referring to the technique of combining similar pixels into a unified region or cluster. Each superpixel \( {P_j} = \bigcup\nolimits_i^{{N_j}} {{p_i}} \) contains adjacent pixels \({{p_i}}\) of \(N_j\) number, similar in color, brightness, texture, or other attributes. SPGQ performs forward differences on three different channels of each superpixel independently. For all pixels within the same channel of each superpixel, the same disturbance is added: \(E_j^c = \sum\limits_i^{{N_j}} {{e_i^c}}  \). The victim model is queried to calculate the difference and obtain an approximate gradient:
\begin{equation}
g_j^c = \frac{{\partial f}}{{\partial P_j^c}} \approx \frac{{f\left( {x + \varepsilon E_j^c} \right) - f\left( x \right)}}{\varepsilon }
  \label{eq:sp difference}
\end{equation} 
Since a superpixel typically contains 900 or more pixels within a channel, perturbing all these pixels will not produce an extremely small pairwise distance, which would not affect the Gaussian distribution of the queried images. Thus, superpixel queries can circumvent Prada attacks. 

\subsection{Sample Gradient Purification}
Sample gradients have a significant variance, and numerical discrepancies between different models can be inherited through the backpropagation process, manifesting in the sample gradients. superpixel gradient also has the same drawbacks. Hence, we have formulated a Sample Gradient Purification Mechanism to mitigate the interference from variance and other extraneous factors.

Concerning the gradient of every queried superpixel channel \(G^c = \bigcup\nolimits_j^J {{g_j^c}}  \), we initially perform a denoising operation. The core objective of denoising is to preserve the extreme values of the sample gradients, eliminating non-extreme gradient values, as the extreme portions encapsulate the focal points of the model. Additionally, due to the divergent implications of positive gradients (indicating facilitation of the loss function) and negative gradients (indicating impediment), it becomes imperative to independently execute denoising operations on the sets of positive and negative gradients. We commence by calculating the extreme values of every channel's positive and negative gradients: \( g_+^c = \max\left(g_j^c \in \{g_j^c | g_j^c \ge 0\}\right) \) and \( g_-^c = \max\left(g_j^c \in \{g_j^c | g_j^c < 0\}\right) \). The filtered gradients are given by:
\begin{align} \label{filter}{G_+^c} =  \{g_j^c | g_j^c > \beta g_+^c\} \quad  \quad {G_-^c} =  \{g_j^c | g_j^c < \beta g_-^c\}\end{align} 
% \begin{equation} {G_-^c} =  \{g_j^c | g_j^c < \beta g_-^c\} \end{equation}
In Equation \ref{filter}, \( \beta \) is a predefined hyperparameter within the range \( 0 < \beta < 1 \), typically set to 0.5 by default. The Discarded gradient set \( {\left( {{G^c}} \right)^\prime } = \left\{ {g_j^c|g_j^c \le \beta g_ + ^c} \right\}\bigcup {\left\{ {g_j^c|g_j^c \ge \beta g_ - ^c} \right\}}   \), is assigned the value of zero. Simultaneously, to avert inheriting numerical discrepancies from the models and potential exaggeration or diminution of gradient values due to gradient explosion or vanishing, normalization of the filtered sample gradients is requisite. The normalization operation is as follows: 
\begin{align} G_ + ^c = \left\{ {\frac{{g_j^c}}{{g_ + ^c}}|g_j^c \in G_ + ^c} \right\}  \quad  \quad G_ - ^c = \left\{ {\frac{{g_j^c}}{{g_ - ^c}}|g_j^c \in G_ - ^c} \right\}\end{align}
% \begin{equation} G_ - ^c = \left\{ {\frac{{g_j^c}}{{g_ - ^c}}|g_j^c \in G_ - ^c} \right\} \end{equation}
The gradient values closer to 1 signify a higher degree of facilitation or impediment of the model's loss function at that particular pixel position. Following denoising and normalization, we obtain the purified sample gradients.

The role of SGP is crucial. We demonstrate the effectiveness of SGP's purification in the subsequent ablation study Section \ref{ablation}.

\subsection{Objective Function for Training}
After obtaining the superpixel level gradient of an image in the victim model \(f_v\), the proxy model 
 \(f_s\) cannot directly mimic and learn the dark knowledge of the superpixel gradient. Firstly, we use a function similar to the one used for training the victim model to calculate the sample pixel gradients of the proxy model. Regardless of whether the query result is hard labels or probabilities, we use a cross-entropy function similar to that used in training the victim model to calculate the sample gradients for the proxy model :
\begin{equation}\label{sg-f} f(y, p) = - \log(y_{p}) \end{equation}
For equation \ref{sg-f}, \(y\) is the proxy model's K-dimensional vector output. \(p\) is the victim model's predicted label. Therefore, for the pixel-level gradients \(Q = \bigcup\nolimits_j^J {\bigcup\nolimits_i^{{N_j}} {{q_i}} } = {{\partial f\left( {y,p} \right)} \mathord{\left/
 {\vphantom {{\partial f\left( {y,p} \right)} \partial }} \right.
 \kern-\nulldelimiterspace} \partial }x \) obtained through backpropagation in the white-box proxy model, we adopt a mean coverage method to obtain the same format of the simulated-superpixel gradient. Specifically, based on the superpixel partition of the victim model, we take the mean of all pixel gradients within the same superpixel in the proxy model, replace the original pixel gradients, and obtain the simulated superpixel gradient: 
\begin{equation} {Q'} = \bigcup\nolimits_j^J {\bigcup\nolimits_c^3 {\left\{ {\frac{{\sum\nolimits_i^{{N_j}} {q_i^c} }}{{{N_j}}}|p_i^c \in {P_j}} \right\}} } \end{equation}

Our ultimate objective function or loss function consists of \(L_{grad}\) and initial \(L_{prob}\). For hard-label query mode (giving top-1 probability \(\hat y\) and predicted label p) and probability query mode (giving K-dimensional probability \(\hat y\)), \(L_{prob}\) is different:
\begin{equation}
 {L_{prob}}\left( {y,\hat y} \right) = 
\begin{cases}
 - y_p \bullet \log \left( {\hat y} \right) - \log(y_{p})  \text{if of hard-label }  \\
- \sum\nolimits_k^K {\left( {{y_k}} \right)}  \bullet \log \left( {{{\hat y}_k}} \right)  \text{if of probability}
\end{cases}
\end{equation}
Our gradient loss function \(L_{grad}\) is set in two parts: the first part ensures similarity in the gradient values for each superpixel, and the second part ensures similarity in the overall gradient of the sample. For the first part, \(f_v\) and  \(f_s\) having similar gradients for each superpixel is equivalent to the query results of the image with the perturbed superpixel being similar. Define \(x_j^c = x + \varepsilon E_j^c\) and \(G_{all}^c = G_ + ^c\bigcup {G_ - ^c} \). Then, we can get:
\begin{equation}L_{grad,1} = \sum\nolimits_j^J {\sum\nolimits_c^3 {\left( {{L_{prob}}\left( {{f_s}\left( {x_j^c} \right),{f_v}\left( {x_j^c} \right)} \right)|g_j^c \in G_{all}^c} \right)} } \end{equation}
For the second part, we need to calculate the cosine similarity between G and \({Q'}\):
\begin{equation}L_{grad,2} = 1 - \cos \left( {G,{Q'}} \right)\end{equation}

Based on \({L_{grad}} = L_{grad,1} + L_{grad,2}\), the final loss function of the victim model is:
\begin{equation}L = {L_{prob}}\left( {{f_s}\left( x \right),{f_v}\left( x \right)} \right) + {L_{grad}}\end{equation}

Through this objective function, the proxy model can effectively mimic the SG knowledge of the victim model. We document the visual changes of the simulated superpixel SG of the proxy model during the training process in Supplementary Material.

%% file: sec/4_Experiment.tex
\section{Experiment}
\label{exp}
\begin{table*}[htbp]
\centering
\caption{
The agreement (in \%), test accuracy (in \%), and queries of each method with querying probability or hard label. For our model, we report the average result as well as the standard deviation computed over 10 runs. (\textbf{Boldface}: the best value.)}
\resizebox{0.9\textwidth}{!}{
\begin{tabular}{lccccccccc}
\hline
\multirow{2}{*}{Method (probability)}    & \multicolumn{3}{c}{CUB200 (10k)}         & \multicolumn{3}{c}{CUB200 (15k)}            & \multicolumn{3}{c}{CUB200 (20k)}\\
\cline{2-10}
                           & Agreement        & Acc    &Queries           & Agreement        & Acc      &Queries        & Agreement        & Acc       &Queries       \\
\hline
ZSDB3KD       & 51.47          & 49.32          & 1202k          & 52.31          & 50.67          & 1098k  & 55.33          & 53.08   &1021k\\
DFMS      & 55.21          & 53.19          & 1021k          & 57.41          & 55.36          & 1003k  & 58.44          & 55.98    &1007k\\
EDFBA          & 55.61          & 53.12          & 459k          & 57.67          & 55.69          & 451k  & 58.12          & 55.72     &451k\\
DS & 54.97          & 53.98          & 1077k          & 57.86          & 55.71          & 1021k  & 57.04          & 56.39     &997k\\
DFME                      & 52.31 & 50.17 & 1489k & 53.01 & 51.27 & 1344k  &  55.27    & 53.18    &1311k\\
SPSG(Ours)              & \textbf{60.71}$\pm$0.51 & \textbf{55.47}$\pm$0.23 & 132k$\pm$0.01k & \textbf{65.98}$\pm$0.34 & \textbf{59.34}$\pm$0.52 & 195k$\pm$0.01k & \textbf{67.48}$\pm$0.21          & \textbf{61.21}$\pm$0.11   &271k$\pm$0.01k\\
\hline
\multirow{2}{*}{Method (probability)}    & \multicolumn{3}{c}{Indoor(10k)}         & \multicolumn{3}{c}{Indoor(15k)}            & \multicolumn{3}{c}{Indoor(20k)}\\
\cline{2-10}
                           & Agreement        & Acc    &Queries           & Agreement        & Acc      &Queries        & Agreement        & Acc       &Queries       \\
\hline
ZSDB3KD       & 59.41          & 58.37          & 1112k          & 63.61          & 63.07          & 988k  & 67.13          & 65.08   &977k\\
DFMS      & 61.11          & 60.12          & 1009k          & 64.58          & 62.26          & 993k  & 68.14          & 67.18    &972k\\
EDFBA          & \textbf{61.17}          & \textbf{60.42}          & 349k          & 64.17          & 62.12          & 311k  & 67.10          & 64.72     &302k\\
DS & 60.27          & 59.98          & 1021k          & 62.12          & 61.78          & 1010k  & 66.04          & 65.91     &982k\\
DFME                     & 58.71 & 58.17 & 1339k & 60.21 & 59.26 & 1294k  &  66.27    & 64.16    &1209k\\
SPSG(Ours)              &58.81$\pm$0.11 & 57.99$\pm$0.13 & 137k$\pm$0.01k & \textbf{64.98}$\pm$0.34 & \textbf{63.34}$\pm$0.12 & 181k$\pm$0.01k & \textbf{70.11}$\pm$0.21          & \textbf{70.27}$\pm$0.11   & 267k$\pm$0.01k\\
\hline
\multirow{2}{*}{Method (probability)}    & \multicolumn{3}{c}{Caltech256 (10k)}         & \multicolumn{3}{c}{Caltech256 (15k)}            & \multicolumn{3}{c}{Caltehch256 (20k)}\\
\cline{2-10}
                           & Agreement        & Acc    &Queries           & Agreement        & Acc      &Queries        & Agreement        & Acc       &Queries       \\
\hline
knockoff       & 51.47          & 49.32          & 10k          & 52.31          & 50.67          & 15k  & 55.33          & 53.08   &20k\\
ActiveThief     & 55.21          & 53.19          & 10k          & 57.41          & 55.36          & 15k  & 58.44          & 55.98    &20k\\
Black-Box Dissector          & 55.61          & 53.12          & 150k          & 57.67          & 55.69          & 220k  & 58.12          & 55.72     &300k\\
InverseNet            & 56.19          & 55.32          & 150k          & 57.79          & 55.82          & 220k  & 58.73          & 56.74     &300k\\
SPSG(Ours)              & \textbf{60.71}$\pm$0.51 & \textbf{55.47}$\pm$0.23 & 132k$\pm$0.01k & \textbf{65.98}$\pm$0.34 & \textbf{59.34}$\pm$0.52 & 185k$\pm$0.01k & \textbf{70.21}$\pm$0.21          & \textbf{63.21}$\pm$0.11   &271k$\pm$0.01k\\
\hline
\multirow{2}{*}{Method (probability)}    & \multicolumn{3}{c}{Diabetic5(10k)}         & \multicolumn{3}{c}{Diabetic5(15k)}            & \multicolumn{3}{c}{Diabetic5(20k)}\\
\cline{2-10}
                           & Agreement        & Acc    &Queries           & Agreement        & Acc      &Queries        & Agreement        & Acc       &Queries       \\
\hline
knockoff       & 31.12          & 29.32          & 10k          & 34.43          & 32.17          & 15k  & 39.56          & 38.17   &20k\\
ActiveThief     & 32.21          & 31.19          & 10k          &35.32          & 34.61          & 15k  & 38.14          & 37.18    &20k\\
Black-Box Dissector          & 34.81 & 33.27 & 150k & 36.10 & 36.27 & 220k  &  40.22    & 39.86    &300k\\
InverseNet          & 35.67 & 34.02 & 150k & 36.81 & 36.01 & 220k  &  41.12    & 40.16    &300k\\
SPSG(Ours)              &\textbf{36.12}$\pm$0.31 & \textbf{35.25}$\pm$0.12 & 122k$\pm$0.01k & \textbf{38.13}$\pm$0.27 & \textbf{37.24}$\pm$0.12 & 179k$\pm$0.01k & \textbf{42.14}$\pm$0.21          & \textbf{41.27}$\pm$0.11   & 278k$\pm$0.01k\\
\hline
\multirow{2}{*}{Method (hard-label)}    & \multicolumn{3}{c}{CUB200 (10k)}         & \multicolumn{3}{c}{CUB200 (15k)}            & \multicolumn{3}{c}{CUB200 (20k)}\\
\cline{2-10}
                           & Agreement        & Acc    &Queries           & Agreement        & Acc      &Queries        & Agreement        & Acc       &Queries       \\
\hline
ZSDB3KD       & 24.45          & 23.56          & 1299k          & 26.33          & 26.01          & 1098k  & 31.31          & 29.78   &931k\\
DFMS      & 26.21          & 25.11          & 1331k          & 29.31          & 28.33          & 1003k  & 31.24          & 30.18    &939k\\
EDFBA          & 25.34          & 23.54          & 559k          & 27.61          & 26.43          & 490k  & 31.12          & 30.72     &431k\\
DS & 24.56          & 23.53          & 1237k          & 27.34          & 26.78          & 1191k  & 30.64          & 29.78     &902k\\
knockoff                     & 21.11 & 19.27 & 10k & 24.49 & 22.76 & 15k  &  26.33    & 25.92    &20k\\
ActiveThief                     & 23.21 & 22.89 & 10k & 26.55 & 25.16 & 15k  &  27.18    & 26.96    &20k\\
Black-Box Dissector                     & 25.91 & 23.57 & 150k & 27.43 & 26.26 & 220k  &  31.59    &30.46    &300k\\
InverseNet                    & 26.01 & 24.07 & 150k & 26.93 & 26.12 & 220k  &  31.43    &30.97    &300k\\
SPSG(Ours)              & \textbf{26.78}$\pm$0.16 & \textbf{25.42}$\pm$0.23 & 132k$\pm$0.01k & \textbf{29.97}$\pm$0.34 & \textbf{29.84}$\pm$0.52 & 195k$\pm$0.01k & \textbf{34.66}$\pm$0.21          & \textbf{34.12}$\pm$0.19   &271k$\pm$0.01k\\
\hline
\end{tabular}
}
\label{Real_dataresult}
\end{table*}
\subsection{Experiment Setup}
\textbf{Victim Model.} We employ four datasets used in Knockoff for our experimentation: Caltech256 (256 classes) \cite{caltech256}, CUB-200-2011 (200 classes) \cite{CUB_200_2011}, Indoor Scenes (67 classes) \cite{Indoorscene}, and Diabetic Retinopathy (5 classes) \cite{DiabeticRetinopath}. For Diabetic Retinopathy, we strip 200 images from the training set for each category, forming a test set that in total contains 1000 images. A resnet34 \cite{resnet} model trained on these four datasets serves as our victim model. The training procedure for the victim model mimics that of Knockoff's victim model. Specifically, the model is trained for 200 epochs using an SGD optimizer with a momentum of 0.5 and an initial learning rate of 0.1 that decays by a factor of 0.1 every 60 epochs. The well-trained victim model is available for download at Knockoff Code. The victim model has achieved accuracies of 78.4\%, 77.1\%, 76.0\%, and 59.4\% on the datasets mentioned sequentially above.

\textbf{Baseline and Attack Dataset.} Baselines are categorized based on the necessity of real data. Data-free baselines include DFME \cite{DFME}, DS \cite{DS}, DFMS \cite{DFMS}, EDFBA \cite{EDFBA}, and ZSDB3KD \cite{ZSDB3KD}, while Knockoff \cite{knockoff}, ActiveThief \cite{activethief}, Black-Box Dissector \cite{blackdissector}, and InverseNet \cite{inversenet} are Data-driven baselines that require real data. Notably, DFMS, EDFBA, ZSDB3KD, DS, Black-Box Dissector, inverseNet, ActiveThief, and Knockoff remain functional even when the query results are hard labels. For CUB-200-2011 and Indoor Scenes, we conduct comparisons in data-free MS. Even though data-free MS does not necessitate real data, we employ publicly available, potentially related real images as weak image priors for the generator for fairness in comparison. Specifically, for Indoor Scenes, We use indoor scene images from SUN \cite{SUN}, which are distinct from Indoor Scenes categories. For CUB-200-2011, We utilize bird images whose classes are included in NAbird \cite{NAbird} but not present in CUB. In Caltech256 and Diabetic Retinopathy, comparisons are made for Data-driven MS. Among them, the number of queries for InverseNet and Black-Box Dissector is not determined by the number of real samples. To ensure fairness, we set the query volume for these two methods slightly higher than that of SPSG. ILSVRC-2012 \cite{imagenet} training set, consisting of approximately 12 million images, serves as our attack dataset in Data-driven MS. Moreover, we ensure the consistency of real samples across different methods.

\textbf{Training Paradigm and Evaluation Metrics.} Training paradigms are categorized based on the usage of data generators. Generally speaking, only data-free MS necessitates the use of a generator. Under the generator paradigm, the Generator is configured as BigGAN and trained using Adam with a learning rate of 0.001, $\beta_1$ = 0.5, and $\beta_2$ = 0.999. The batch size of BigGAN is 128. For data-driven MS, proxy models are trained from scratch on the attack dataset using SGD with a momentum of 0.5, a learning rate of 0.01 (decaying by a factor of 0.1 every 60 epochs), 200 epochs, and a batch size of 64. Evaluation is based on the accuracy of the proxy model on the corresponding test set and the similarity in predictions between the proxy and victim models. We also report the success rate of adversarial attacks on the victim model as indicators of the transferability of the proxy model. The superpixel segmentation for our method defaults to quickshift \cite{quickshift}.

\subsection{Experiment Results with Data-free MS}
We evaluate the accuracy and agreement of the proxy models generated by data-free baselines and our algorithm under scenarios with 10k, 15k, and 20k real samples. As illustrated in Table \ref{Real_dataresult}, our method almost outperforms all other MS methods across both metrics. While the performance of data-free MS plateaus with increasing numbers of real samples, our method demonstrates less pronounced diminishing returns. Additionally, we document the query volume required by all algorithms. Our method’s query volume is determined by the number of queries consisting of an image and its perturbed versions using different superpixels, whereas the query volume for data-free MS depends on when the model converges. Our method requires significantly fewer queries than data-free MS, with our query volume being approximately 25$\%$ of that required by the most efficient data-free MS. Notably, despite the use of weak image priors, DFMS and ZSDB3KD sometimes fail to train, severely limiting their practical applicability.

\subsection{Experiment Results with Data-driven MS}
\begin{figure*}[htbp]
    \centering
    \begin{minipage}{0.3\textwidth}
        \includegraphics[width=\linewidth]{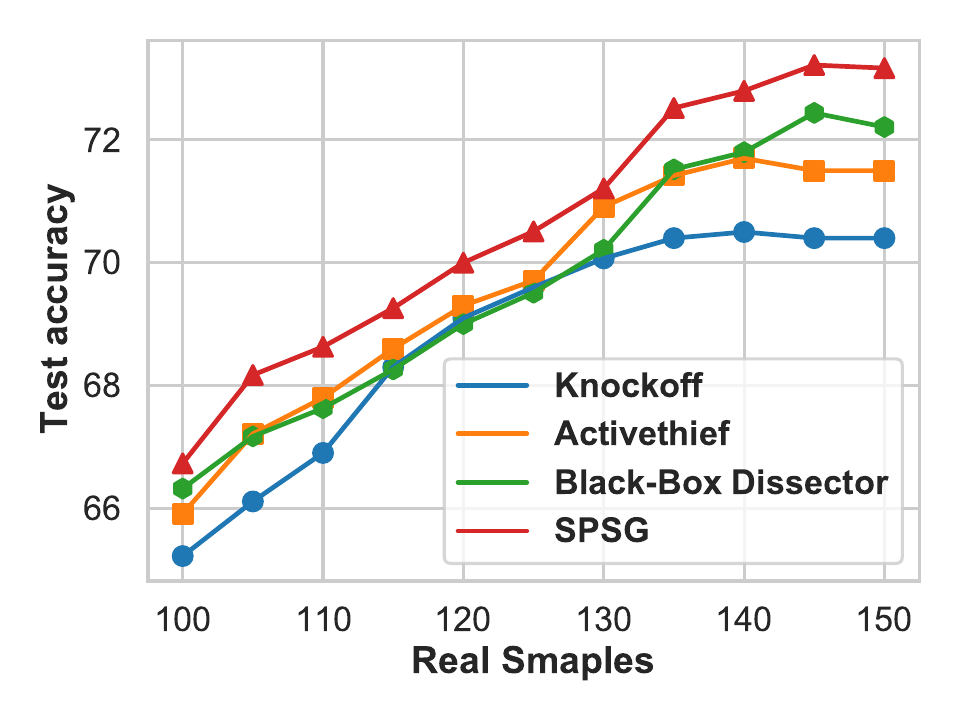}
        \caption{Baselines in CUB200}
        \label{fig:first_image}
    \end{minipage}
    \begin{minipage}{0.3\textwidth} 
        \includegraphics[width=\linewidth]{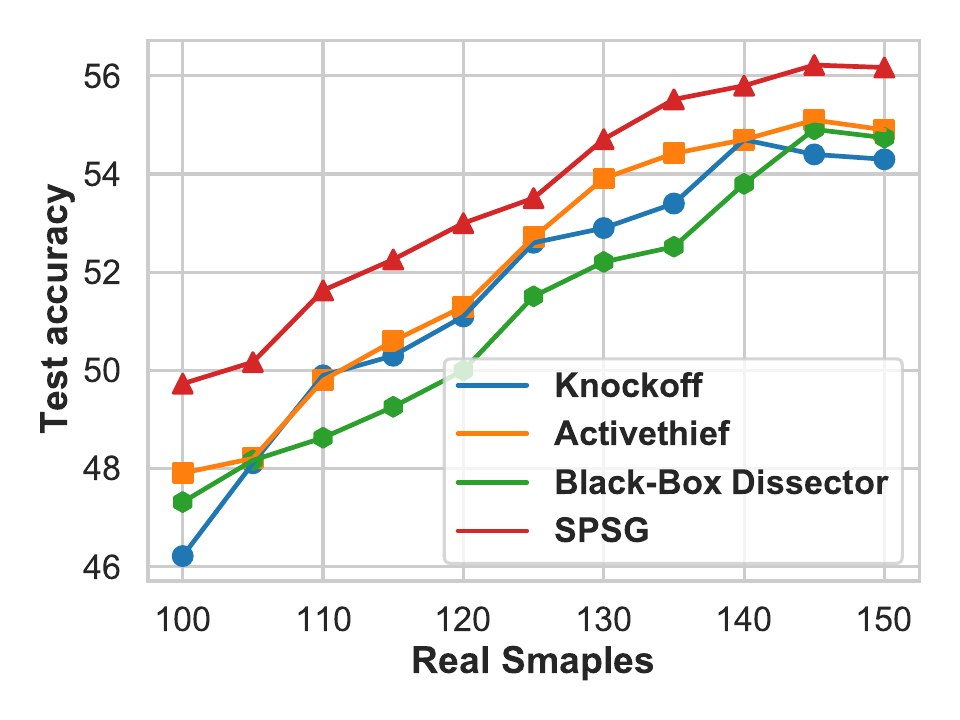}
        \caption{Baselines in Diabetic5}
        \label{fig:second_image}
    \end{minipage}
    \begin{minipage}{0.3\textwidth} 
        \includegraphics[width=\linewidth]{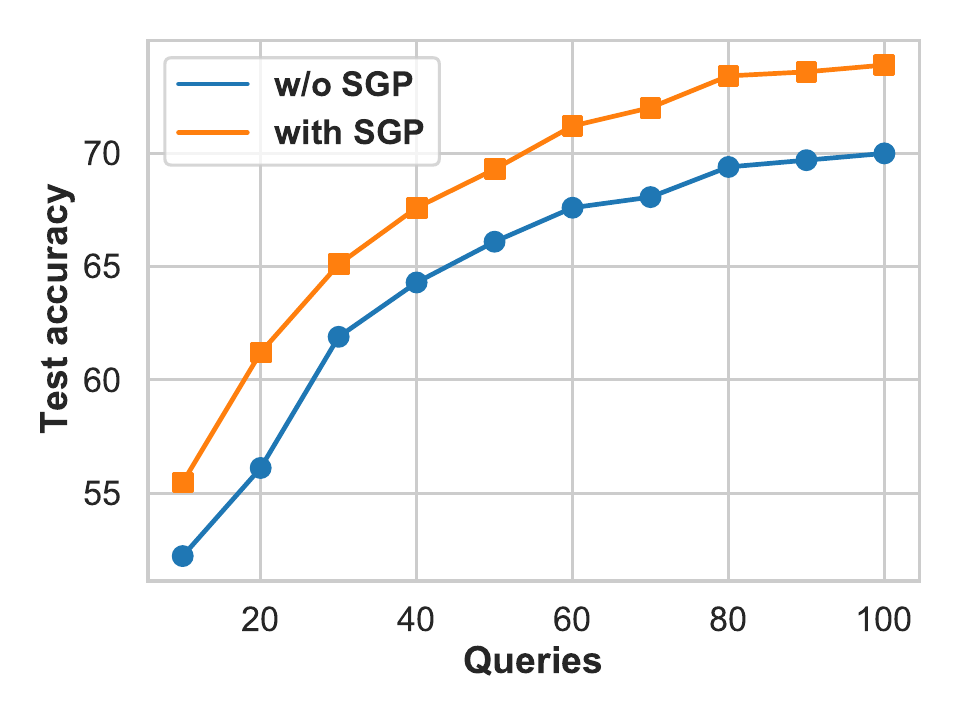}
        \caption{Ablation study}
        \label{fig:third_image}
    \end{minipage}
\end{figure*}

We assess the accuracy and agreement of the proxy models generated by SPSG and other data-driven MS methods under scenarios with 10k, 15k, and 20k real samples. As shown in Table \ref{Real_dataresult}, while our method does not minimize query volume, it yields substantial improvements. The marginal effects of data-driven  MS are not pronounced across the 10k to 20k real samples range. Consequently, we observe the performance across a broader range of real sample numbers (100k-200k), finding that our algorithm consistently outperforms others in terms of accuracy. Specifically, as shown in Figure \ref{fig:first_image} and \ref{fig:second_image}, at 140k real samples for Diabetic Retinopathy, Knockoff achieves its peak accuracy of 54.7$\%$, whereas our algorithm surpasses this accuracy at 130k real samples.

\subsection{Experiment Results in Hard-label Query} 
We evaluate all algorithms functional under hard-label queries, where the query not only returns the predicted label but also the associated confidence. We argue that providing users with the confidence associated with the predicted label is more practical for MLaaS. Despite high confidence not necessarily equating to accuracy, low confidence allows users to disregard the model’s prediction. Table \ref{Real_dataresult} presents our results on CUB-200-2011, showcasing our algorithm’s superiority with 10k or more real samples.

\subsection{Transferability of Adversarial Samples}
We assess the transferability of adversarial samples generated on CUBS-200-2011 test set. The evaluation encompasses the success rates of adversarial attacks generated from different methods (FGSM, BIM, PGDK), with a perturbation bound of $\varepsilon$=10/255 and a step size of $\alpha$=2/255. The adversarial attacks in our experiments are untargeted attacks. In untargeted attacks, adversarial samples are generated only on images correctly classified by the attacked model. %, while in targeted attacks, they are generated only on images not classified into a specific incorrect label.
As Table \ref{Adversarial} demonstrates, the adversarial samples generated by our proxy model exhibit higher transferability to the victim model, affirming the practical applicability of our method in real-world scenarios.
\begin{table}[htbp]
    \centering
    \caption{The ASR(in \%) of MS methods with different adversarial attacks on CUB200. (\textbf{Boldface}: the best value.)}
    \vspace{-0.3cm}
    \resizebox{\columnwidth}{!}{
    \begin{tabular}{lcccccc}
    \hline
    \multirow{2}{*}{Method} & \multicolumn{3}{c}{CUBS200 (hard-label)}          & \multicolumn{3}{c}{CUBS200 (probability)}  \\
    \cline{2-7}
    & FGSM & BIM & PGD & FGSM & BIM & PGDk     \\
    \hline
    ZSDB3KD            & 21.31\%    & 23.62\%    & 23.58\%    & 37.64\% & 37.10\% & 39.21\%  \\
    DFMS   & 24.34\%    & 22.93\%    & 21.22\%    & 35.22\% & 34.33\% & 36.11\%  \\
    EDFBA    & 24.21\%    & 23.72\%    & 25.51\%    & 37.88\% & 39.09\% & 38.51\%  \\
    DS    & 25.21\%    & 28.77\%    & 23.51\%    & 37.88\% & 39.09\% & 39.51\%  \\
    knockoff    & 21.31\%    & 24.77\%    & 22.51\%    & 35.12\% & 36.71\% & 37.57\%  \\
    ActiveThief    & 24.34\%    & 26.77\%    & 27.52\%    & 37.88\% & 36.02\% & 36.29\%  \\
    Black-Box Dissector   & 24.91\% & 26.57\% & 26.41\% & 37.21\%  &  36.11\%    &36.46\%\\
    InverseNet   & 24.86\% & 26.76\% & 27.22\% & 38.03\%  &  37.21\%    &37.47\%\\
    Ours                    & \textbf{25.47\%}    & \textbf{29.03\%}   & \textbf{29.31\%}    & \textbf{38.21\%} & \textbf{39.43\%} & \textbf{39.68\%}  \\
    \hline
    \vspace{-0.5cm}
    \end{tabular}
    }
    \label{Adversarial}
\end{table}

\begin{table}[htbp]
    \centering
    \caption{The agreement (in \%) and test accuracy (in \%) of different segment methods on CUB-200-2011.}
    \vspace{-0.3cm}
    \resizebox{\columnwidth}{!}{
    \begin{tabular}{lcccc}
    \hline
    \multirow{2}{*}{Method} & \multicolumn{2}{c}{CUBS200 (hard-label)}          & \multicolumn{2}{c}{CUBS200 (probability)}  \\
    \cline{2-5}
    & agreements & accuracy  & agreements & accuracy     \\
    \hline
    quickshift (132k)           & 26.78\%    & 25.42\%       & 60.71\% & 55.47\%    \\
    felzenszwalb (1871k)   & 30.34\%    & 28.93\%      & 63.22\% & 59.33\%   \\
    slic (371k)    & 29.21\%    & 27.72\%   & 62.28\% & 58.09\%    \\
    Grid (2700k)   & 12.31\%    & 9.72\%   & 21.68\% & 19.09\%   \\
    \hline
    \vspace{-0.5cm}
    \end{tabular}
    }
    \label{superseg}
\end{table}

\begin{figure}[htbp]
    \centering
    \begin{subfigure}[b]{0.45\columnwidth}
    \includegraphics[width=\textwidth]{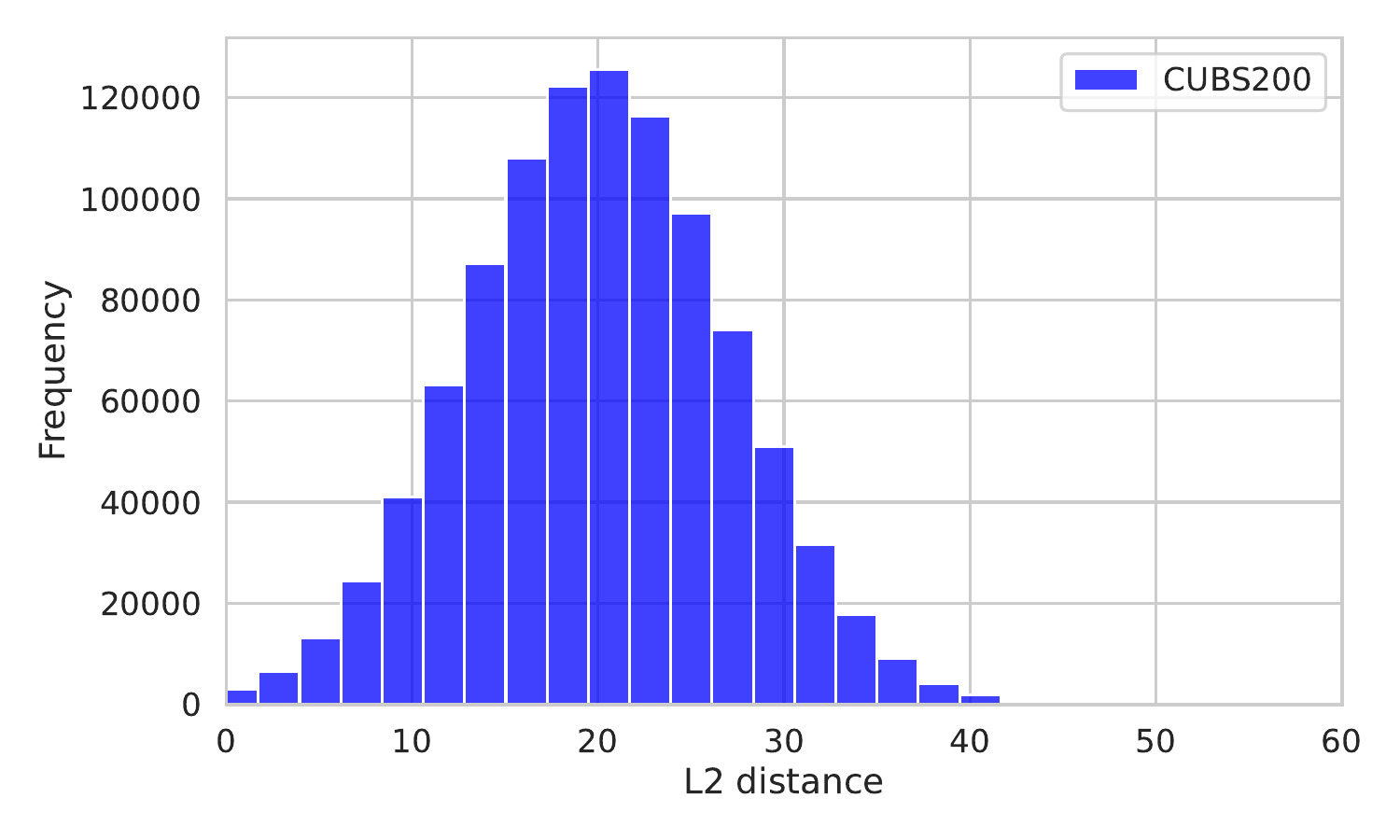}
    \caption{queries for CUB200 model}
    \end{subfigure}
    \begin{subfigure}[b]{0.45\columnwidth}
    \includegraphics[width=\textwidth]{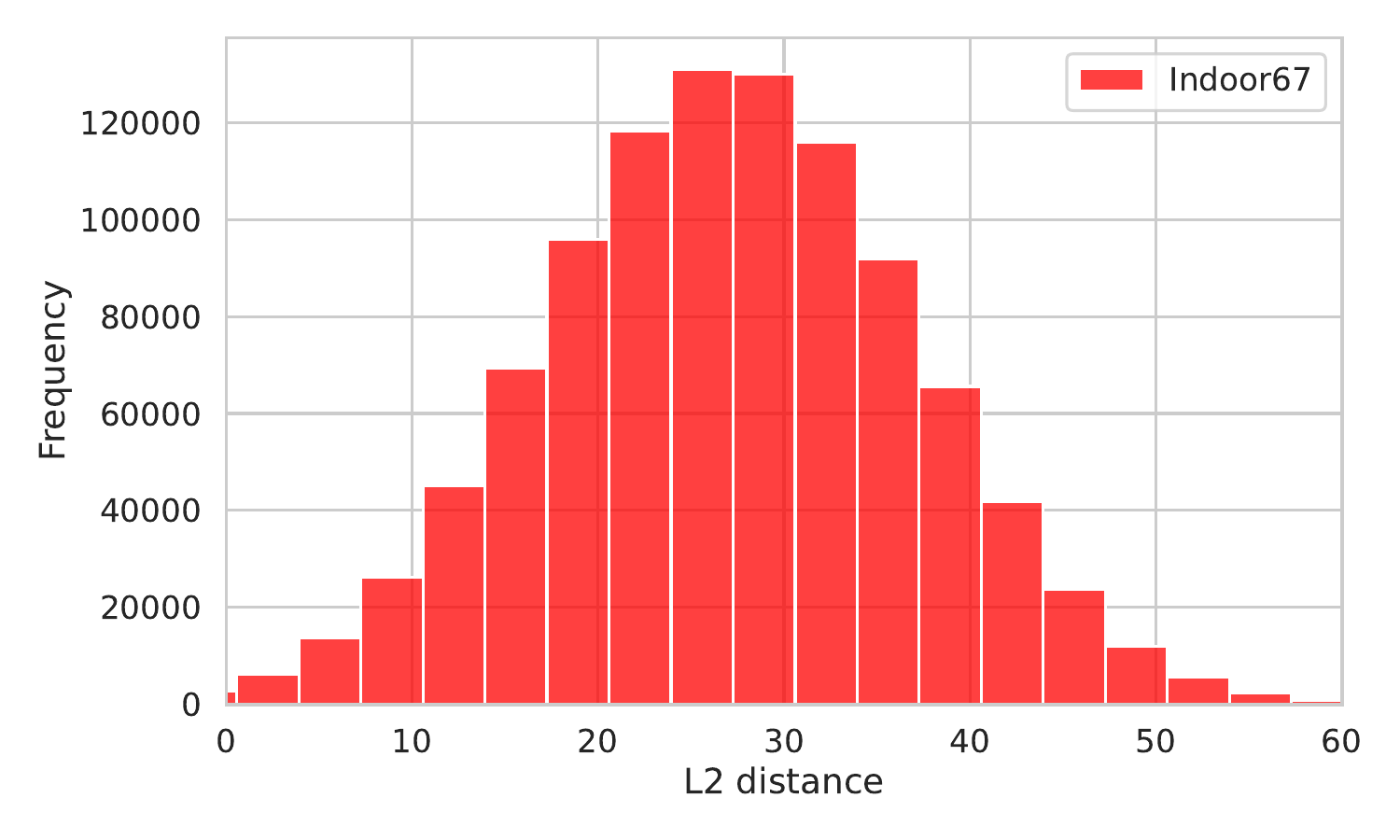}
    \caption{queries for Indoor67 model}
    \end{subfigure}\\
    \begin{subfigure}[b]{0.45\columnwidth}
    \includegraphics[width=\textwidth]{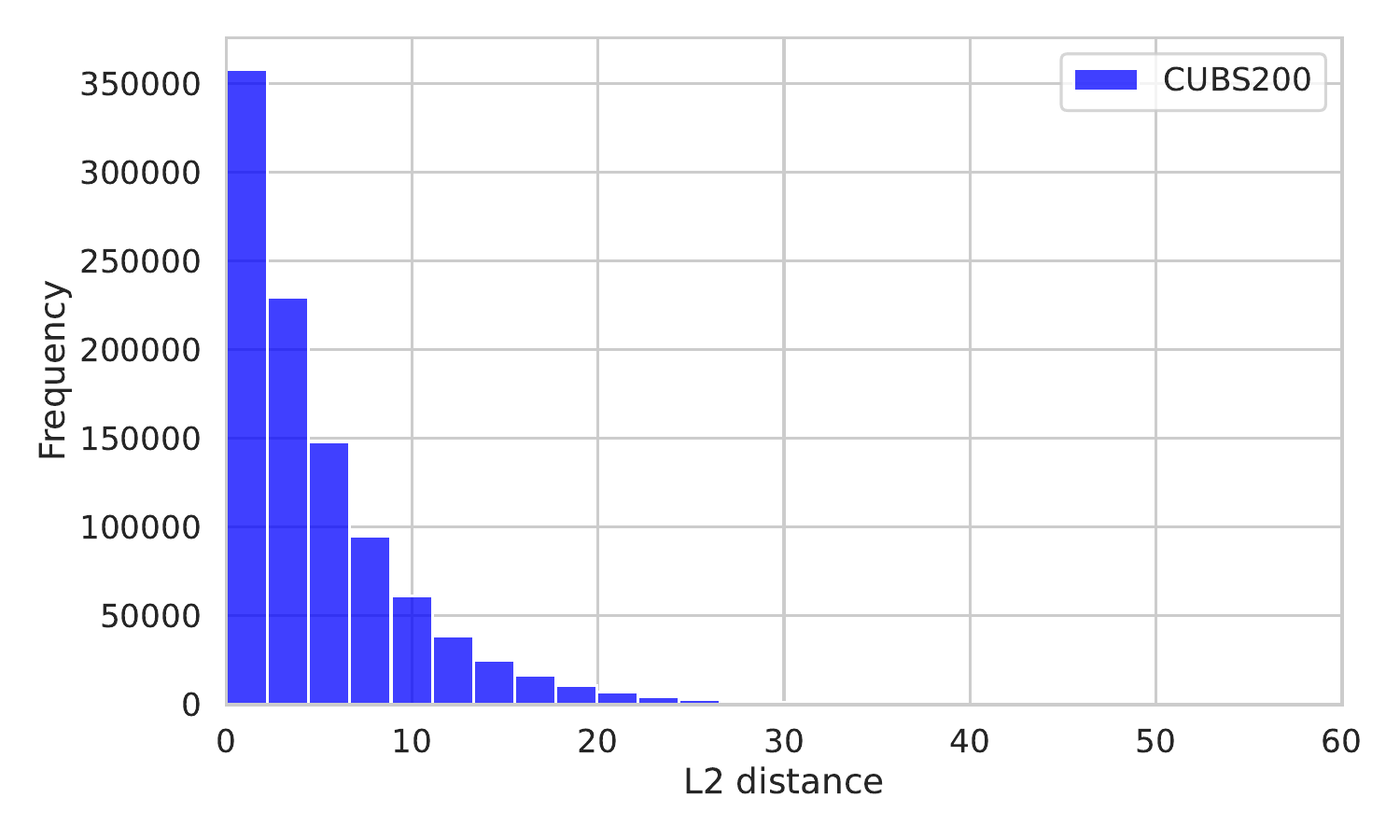}
    \caption{queries for CUB200 model}
    \end{subfigure}
    \begin{subfigure}[b]{0.45\columnwidth}
    \includegraphics[width=\textwidth]{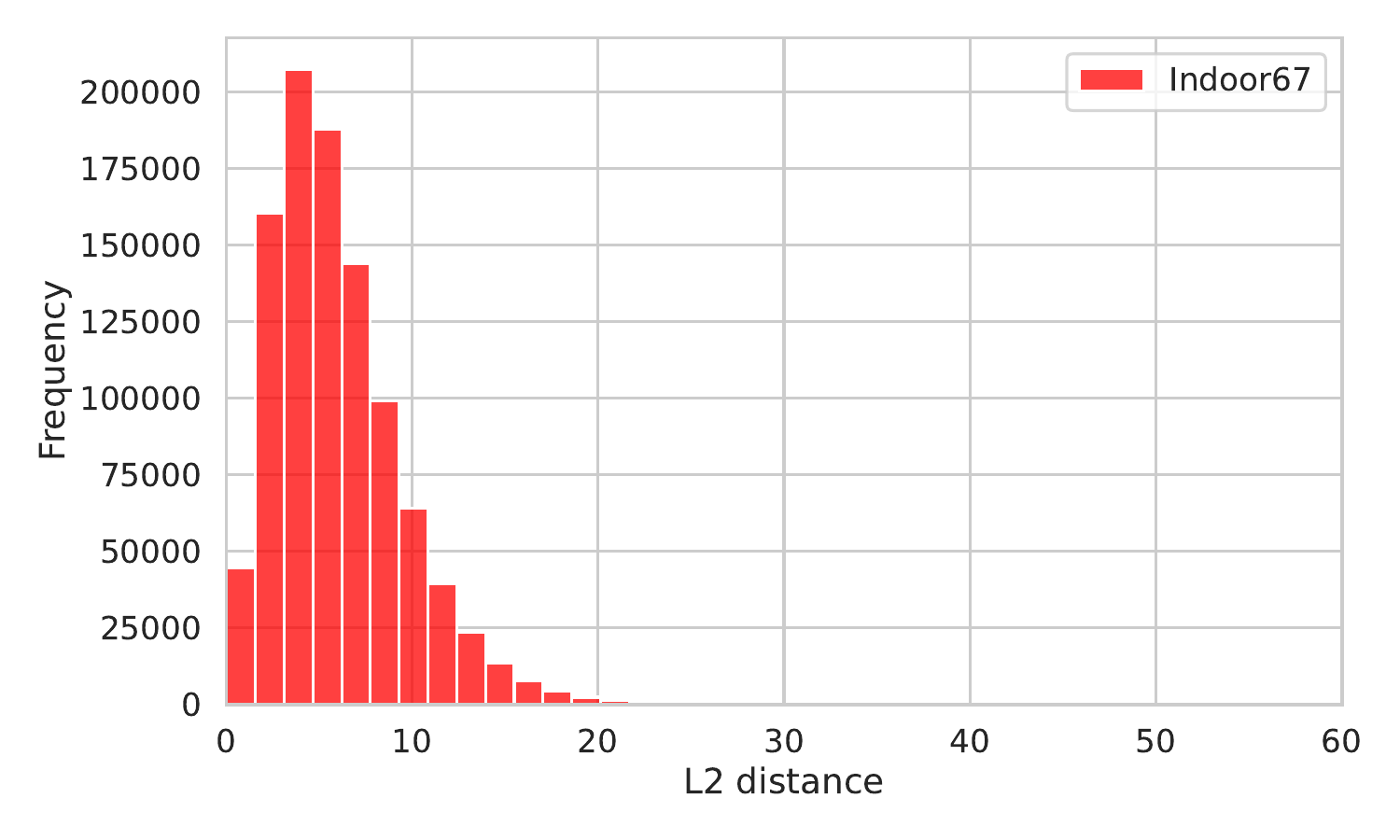}
    \caption{queries for Indoor67 model}
    \end{subfigure}
    \caption{The first and second row show the 1000k queries' image distance distribution for SPGQ and finite difference queries, respectively. }
    \label{fig:ablation}
\end{figure}
\subsection{Model Stealing in Real-World Scenarios}
We trained a model on the Oxford 102 Flowers dataset \cite{Flowers} using the Microsoft Custom Vision \cite{MicrosoftCustomVision} service and designated it as a black-box victim model. The ILSVRC-2012 dataset served as the attack dataset, with the inference results of the Oxford 102 Flowers test set used as the metric. Hyperparameter settings were consistent with previous experiments. The victim model's test accuracy was 86.34\%. As Table \ref{tab:arc} demonstrates, compared to the second-best methods using 30k real samples, our method showed a 4.17\% increase in test accuracy for the proxy model. This result indicates that our method possesses stronger practical applicability in real-world scenarios.
\begin{table}[htbp]
\caption{Test accuracy of all baselines in Real-World Scenarios.}
\vspace{-0.3cm}
    \centering
    \resizebox{\columnwidth}{!}{
    \begin{tabular}{lccccc}
    \hline
    \multirow{2}{*}{Method (probability)} & \multicolumn{5}{c}{Real Sample Number}\\
    \cline{2-6}
                            & 10k & 15k & 20k & 25k & 30k      \\
    \hline
    KnockoffNets            & 58.45\%    & 62.12\%    & 66.48\%    & 69.12\% & 74.50\%   \\
    ActiveThief             & 60.90\%    & 64.21\%    & 68.51\%    & 70.48\% & 75.24\% \\
    Black-Box Dissector     & 60.21\%    & 64.27\%    & 67.52\%    & 69.88\% & 73.09\% \\
    SPSG                    & \textbf{62.49\%}    & \textbf{65.93\%}    & \textbf{69.34\%}    & \textbf{71.37\%} & \textbf{79.41\%}   \\
    \hline
    \end{tabular}
    }
    \label{tab:arc}
    \vspace{-0.5cm}
\end{table}

\subsection{Ablation Study}\label{ablation}
\textbf{Resistance to Prada.} We document the monitoring of the finite difference query and SPGQ by Prada, represented by the distribution of image distances. As shown in Figure \ref{fig:ablation}, Finite difference queries are completely detectable by Prada, exhibiting a significant deviation from the Gaussian distribution. In contrast, superpixel queries initially have a few detectable instances but subsequently evade detection entirely, with their image distribution closely aligning with the Gaussian distribution.
\begin{figure}[htbp]
    \vspace{-0.2cm}
    \centering
    \includegraphics[width=1.\columnwidth]{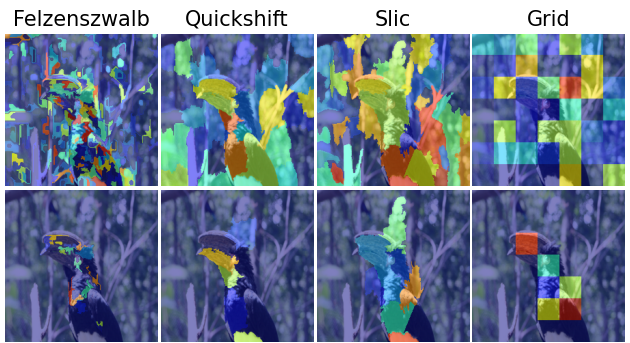}
    \caption{The first row shows the superpixel SG heatmap for different segment methods, while the second row shows the purified superpixel SG map for different segment methods.}
    \vspace{-0.5cm}
    \label{fig:segSG}
\end{figure}

\textbf{Superpixel Segmentation.} We employed the superpixel gradients obtained through queries under quick-shift \cite{quickshift}, felzenszwalb \cite{felzenszwalb}, slic \cite{slic}, and grid segmentation methods. Experimental results in Table \ref{superseg} indicate that the more superpixels used, the more apparent the effect of model stealing becomes. However, regardless of the number of grid pixels divided, the grid segmentation method demonstrated a poor stealing effect. This is attributed to the grid segmentation's disregard for image attributes such as texture and color, which are closely associated with the model's decision-making process.

\textbf{Impact of SGP.} We conduct experiments on CUB-200-2011 dataset to compare the performance of SPSG without SGP and the complete SPSG. The experimental results shown in Figure \ref{fig:third_image} reveal a significant degradation in the effectiveness of SPSG when SGP is omitted. This decline can be attributed to the retention of gradient variance, which proves to be particularly detrimental to dark knowledge extraction. In Supplementary Material, we further explore the applications of SGP, including knowledge distillation.

%% file: sec/5_conclusion.tex
\section{Conclusion}
SPSG significantly outperforms existing MS algorithms across various datasets, demonstrating its effectiveness even in hard-label query scenarios. The success of SPSG in adversarial attacks showcases its practical utility, while its capability to evade Prada highlights its stealthiness. In essence, SPSG provides a novel approach to enhancing MS performance by effectively mimicking additional information from victim models. We hope our proposed method will encourage proactive measures to protect models against unauthorized access and theft.\\
\textbf{Acknowledgments} This work was supported by the National Natural Science Foundation of China Project (62172449, 62372471, 62172441), the Joint Funds for Railway Fundamental Research of National Natural Science Foundation of China (Grant No. U2368201), special fund of National Key Laboratory of Ni\&Co Associated Minerals Resources Development and Comprehensive Utilization(GZSYS-KY-2022-018, GZSYS-KY-2022-024), Key Project of Shenzhen City Special Fund for Fundamental Research(JCYJ20220818103200002), the National Natural Science Foundation of Hunan Province(2023JJ30696), and the Science Foundation for Distinguished Young Scholars of Hunan Province (NO. 2023JJ10080).

%% file: sec/X_suppl.tex
\maketitlesupplementary
\setcounter{page}{1}
\section{Why does Sample Gradient contain model information?}
\begin{figure*}[h]
    \centering
    \includegraphics[width=0.8\textwidth]{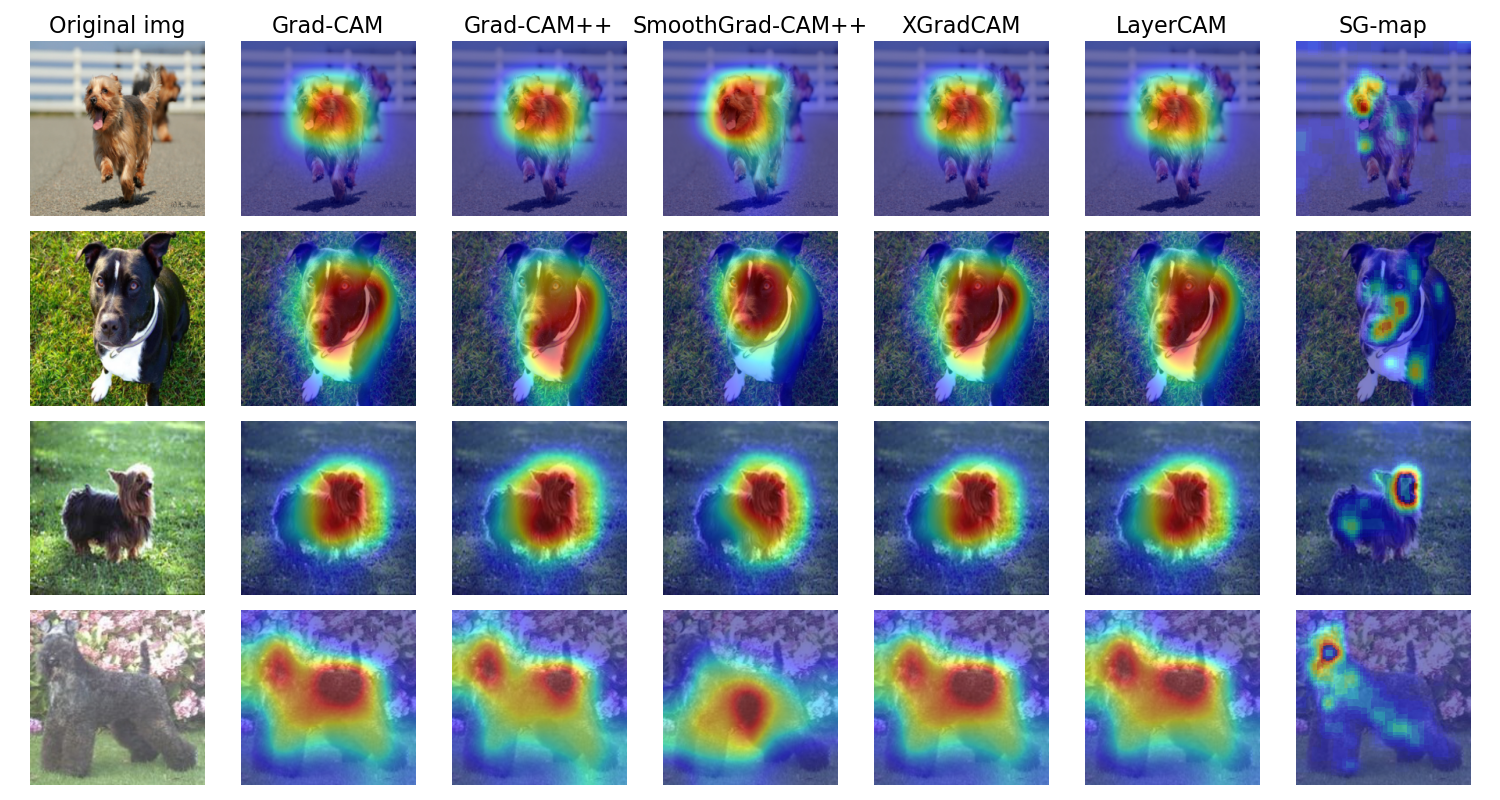}
    \caption{Partial visualization results of grad-CAM \cite{gradCAM}, grad-CAM++ \cite{gradcam++}, Smooth-gradCAM \cite{smoothgradcam++}, X-gradCAM \cite{xcam}, layer-CAM \cite{layercam}, and SG-map. The neural network is ResNet34 pre-trained on ILSVRC-2012.}
    \label{SG-map}
\end{figure*}
\label{sec:formatting}
'Dark knowledge' is a term used to describe knowledge that is implicitly embedded in a model but doesn't manifest directly, such as predictive logits. By learning 'dark knowledge', a surrogate model can inherit the characteristics and mimic the functionalities of the victim model. In prior work, sample gradients have been interpreted as a reflection of a model's local sensitivity to a specific input, guiding the perturbation direction in adversarial attacks. However, due to the inclusion of variance, instability, and a lack of interpretability, sample gradients are rarely considered as a form of dark knowledge. Numerous studies \cite{CAM,gradCAM,GuidedBackpropagationg,DeepLIFT,Smoothgrad,IntergradedG,sliencemap} have attempted to utilize sample gradients to aid interpretability, often through altering the model, employing the gradient of the model's feature maps, or introducing additional inputs to propose interpretability methods. These methods fall short of establishing the interpretability of the original sample gradients. In this section, we introduce SG-Map, a method for interpreting sample gradients, designed without modifying the original model architecture, utilizing feature maps, or adding any additional inputs. We demonstrate that the sample gradients, processed and visualized as heatmaps, exhibit interpretability comparable to Grad-CAM.

For an individual input image, after backpropagation through the loss function, each pixel is assigned a gradient value, collectively forming the sample gradient. The SG-Map algorithm initiates by preprocessing the sample gradients: taking the absolute value of the gradient for each pixel and normalizing the pixels in each channel independently. This preprocessing ensures that the sample gradients meet the requirements for image display and eliminates numerical discrepancies in the sample gradients, which are tied to the parameter values of all neurons in the model and do not accurately reflect the model's decision-making characteristics. Channel-wise normalization is preferred over whole-image normalization due to the more substantial inter-pixel connections within channels than between them. SG-Map then combines the pixel gradients from the three different channels according to the specifications for a grayscale image, resulting in a single-channel sample gradient. In a crucial final step, we apply average pooling to this single-channel sample gradient, mitigating the impact of erratic behaviors from specific instances of the model on the pixel gradients. The resulting sample gradient is then presented as a heatmap. The visualization result is shown in Figure \ref{SG-map}. Comparing SG-Map with CAM methods, we observe that SG-Map focuses on similar pixel locations, reflecting the model's sensitivity and attention allocation across different areas. Unlike CAM methods, the visualization of sample gradients through SG-Map provides a more stringent expression of sensitivity, manifesting as more concentrated yet precise high-temperature areas in the heatmap. Our proposed SG-Map thereby conclusively demonstrates that sample gradients encapsulate deep-seated information of the model, qualifying as a form of dark knowledge that can guide the training of surrogate models.
\section{More Experiment Result}
\subsection{Hard-label experiments of Indoor-Scene }
Under the query setup with hard labels, we use baselines to steal the resnet34 model trained on Indoor-Scenes. As shown in Table \ref{H_dataresult}, SPSG still maintains the highest performance metrics.
\begin{figure}[htbp]
    \centering
    \begin{subfigure}[b]{0.85\columnwidth}
    \includegraphics[width=\textwidth]{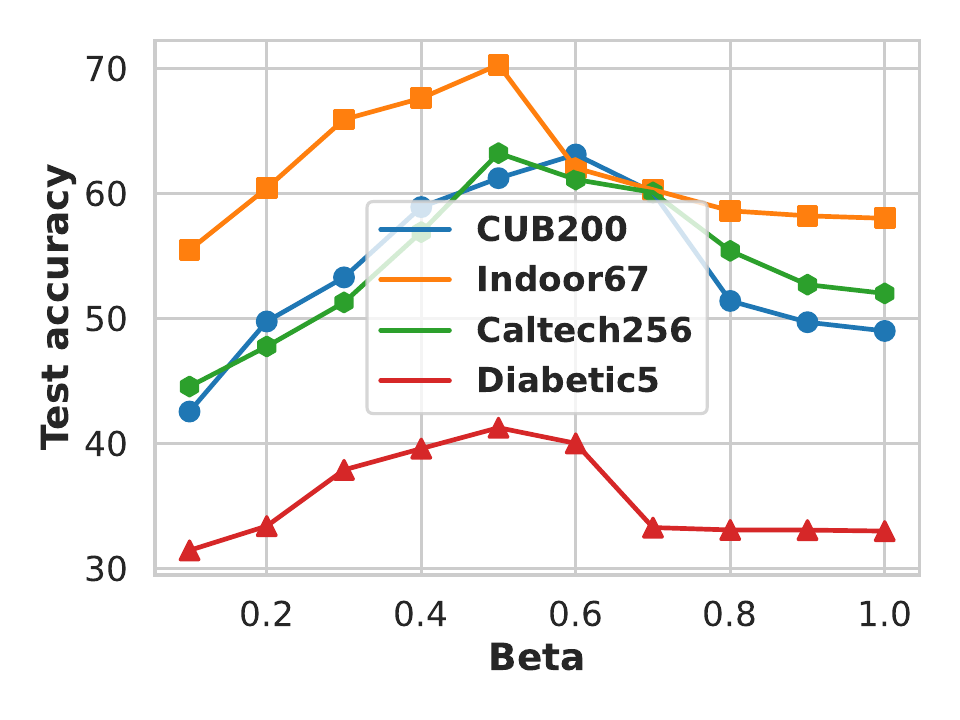}
    \end{subfigure}\\
    \begin{subfigure}[b]{0.85\columnwidth}
    \includegraphics[width=\textwidth]{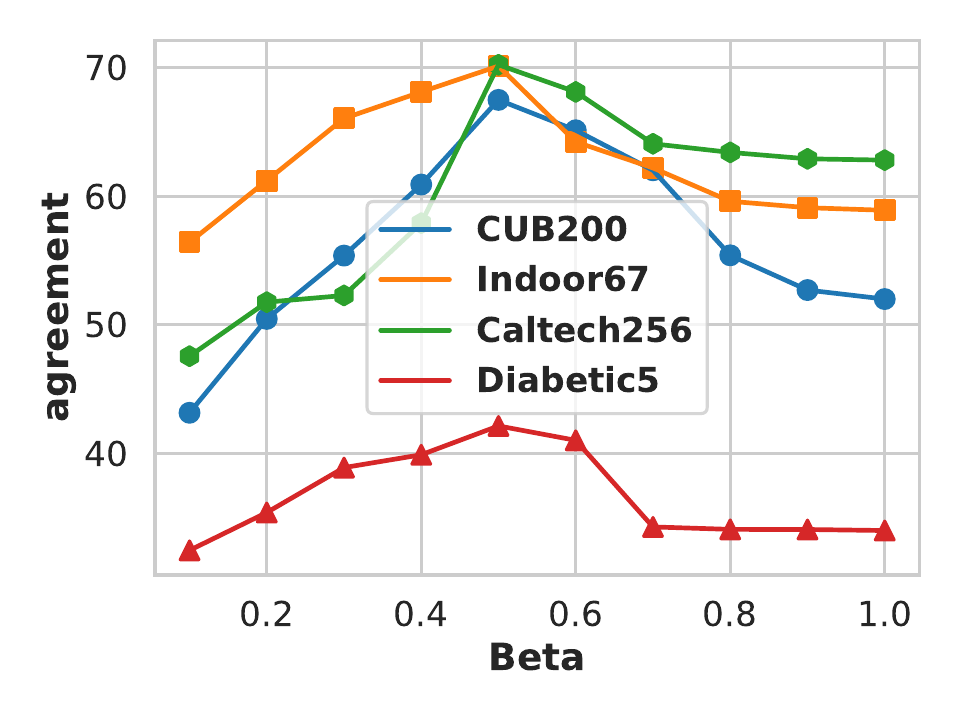}
    \end{subfigure}
    \caption{}
    \label{beta}
\end{figure}

\subsection{Experiments on different Proxy’s architecture }
In the main text, we default to using the same proxy architecture as the victim, that is, resnet34. However, in practice, we cannot obtain information about the victim's model. Therefore, we experiment with different neural network architectures as proxys. The experimental results, as shown in Table \ref{a_dataresult}, indicate that SPSG can effectively extract the performance of the victim across various neural network architectures. Due to the different performance ceilings inherent to each neural network architecture, the results present varying degrees of difference.

\subsection{Impact of hyperparameter $\beta$}
 \(\beta\) is used to remove gradients outside of extreme values. The larger the  \(\beta\), the more gradients are removed, indicating a stricter selection of extremes. We conduct experiments under different  \(\beta\) values. When the value of \(\beta\) is small, more gradient variance is introduced, leading to a decrease in the proxy model's performance. When  \(\beta\) value is larger, there are hardly any superpixel gradients left. When  \(\beta\) is at its maximum, the sample gradient contains only one superpixel gradient. When  \(\beta\) is greater than 0.8, the performance of the proxy model remains at a lower level. This is because at this point, what is left are the most representative superpixel gradients, so the performance of the proxy model remains unchanged as it always mimics the most important superpixel gradients. when \(\beta\) is 0.5, SPSG gets the best performance. The visulaization result is shown in Figure \ref{beta}

\subsection{Offline Training of SPSG}
\begin{figure*}[htbp]
    \centering
    \includegraphics[width=0.9\textwidth]{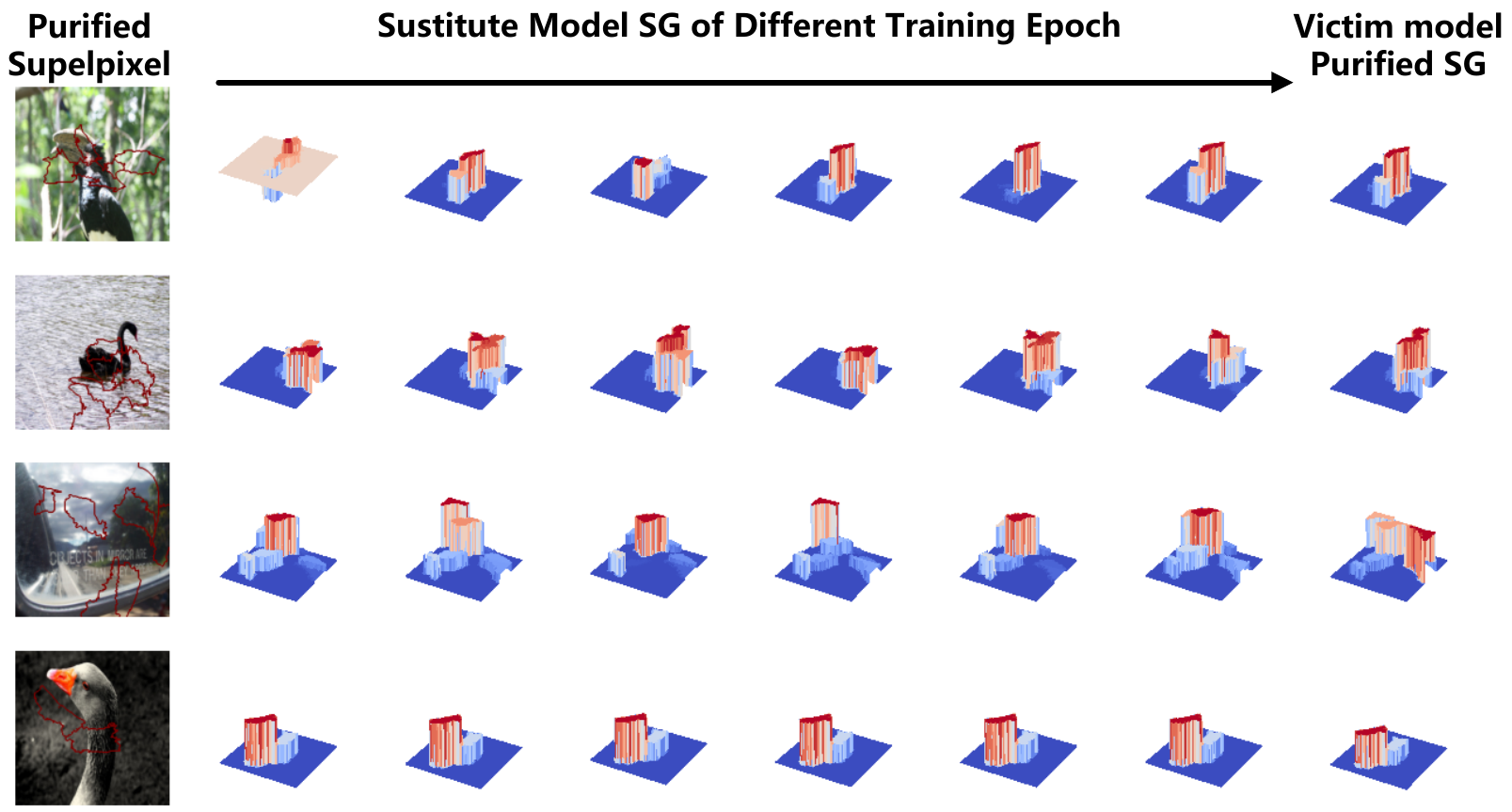}
    \caption{Simulated-superpixel gradients of samples during the offline training process}
    \label{SG-Train}
\end{figure*}

We observe the changes in simulated-superpixel gradients of samples during the offline training process of the proxy model, as shown in Figure \ref{SG-Train}. As the training epochs increase, the similarity between the pseudo-superpixel gradients obtained from the mean of model backpropagation pixel SG and the superpixel gradients queried from the victim model gets higher and higher. More similar sample gradients indicate that our loss function setting is reasonable. The proxy model sufficiently learns SG knowledge of the victim model.

\subsection{Impact of Sample Selection Strategy}
Our method does not conflict with sample selection strategies. Therefore, we compare the performance of SPSG, knockoff, and Black Dissector under two different sample selection strategies. These two strategies are the Reinforcement learning strategy \cite{knockoff} and the K-center strategy \cite{activethief}, as shown in Table \ref{sta_dataresult}. Both strategies improve the performance of MS to varying degrees. SPSG achieves the highest accuracy and similarity under both sample selection strategies. It is important to note that in the main text, we have already found that SPSG also obtains the best performance compared to other methods under a random sample selection strategy.

\subsection{Ability to evade the SOTA defense method.}
We conducted experiments with protection measures similar to \cite{blackdissector} including Adaptive Misinformation \cite{misinformation}, Prediction Poisoning \cite{Predictionpoisoning}, Gradient Redirection\cite{GradientRedirection}, External Feature\cite{ExternalFeature}. The victim model is ResNet34 model trained on the CUBS-200-2011 dataset. The real sample number is 20k. The attack set employed ILSVRC-2012. As shown in Table \ref{tab:mis}, SPSG demonstrates significantly higher resistance to these two defenses compared to other methods. 
\begin{table}[htbp]
\centering
\vspace{-0.3cm}
\caption{
The larger the threshold, the better the defense effect (0.0 means no defense).
"False" and "True" respectively correspond to evading monitoring and being detected by monitoring.
}\label{tab:mis}\vspace{-10pt}
\resizebox{\columnwidth}{!}{
\begin{tabular}{lccccc}
\hline
Method & No defence &AM   &GR  &EF  &PP   \\
\hline
Threshold                        & -                & 0.5    & -    & -  & 0.5\\
\hline
KnockoffNets                                & 54.21$\pm$0.11           & 49.13$\pm$0.28  & 47.17$\pm$0.13 & True & 49.22$\pm$0.11  \\
ActiveThief                      & 55.24$\pm$0.12           & 50.12$\pm$0.11  & 47.71$\pm$0.21  & True & 49.11$\pm$0.21  \\
Black-Box Dissector                        & 56.98$\pm$0.21           & 51.21$\pm$0.31  & 48.81$\pm$0.14  & False & 49.03$\pm$0.28   \\
Inversenet                        & 55.17$\pm$0.19           & 49.21$\pm$0.73  & 51.28$\pm$0.31  & False & 48.72$\pm$0.49   \\
DFMS                       & 52.17$\pm$0.32           & 53.27$\pm$0.21  & 47.79$\pm$0.32  & False & 49.92$\pm$0.09   \\
EDFBA                        & 55.32$\pm$0.21           & 43.22$\pm$0.13  & 49.82$\pm$0.34  & False & 48.87$\pm$0.73   \\
DS                        & 51.33$\pm$0.12           & 54.33$\pm$0.37  & 51.11$\pm$0.94  & True & 50.01$\pm$0.56   \\
DFME                        & 53.28$\pm$0.23           & 52.26$\pm$0.61  & 50.87$\pm$0.34  & False & 51.06$\pm$0.72   \\
SPSG        & \textbf{61.33 }$\pm$0.03 & \textbf{54.95 }$\pm$0.22 & \textbf{52.15 }$\pm$0.34 & False & \textbf{59.02 }$\pm$0.41  \\
\hline
victim model                                & 77.10           & 71.29  & 74.16 & -  & 71.46\\
\hline
\vspace{-0.8cm}
\end{tabular}}
\end{table}

\subsection{More study about SPGQ}
We investigate the effectiveness of SPGQ and finite difference query methods. We compare the similarity of the sample gradients obtained by different methods to the real sample gradients. We use the average pair-wise distance to each real sample gradient as the evaluation metric and record the average number of queries required to query a sample. For superpixel gradients, we compare them by averaging the real samples within the corresponding superpixels. As shown in Table \ref{tab:sg}, our method exhibits a smaller distance compared to the finite difference method, indicating a higher similarity.

\begin{table*}[htbp]
    \centering
    \resizebox{\textwidth}{!}{
    % \begin{tabular}{l|c|c|c|c}
    \begin{tabular}{lccccc}
    \hline
    \multirow{2}{*}{Result} & \multicolumn{5}{c}{Gradient Query Method}\\
    \cline{2-6}
                            & Finite Difference & SPGQ(QuickShift) & SPGQ(Slic)  &SPGQ(Felzenszwalb) &Grid Query \\
    \hline
    Distance            & 2.987    & 2.178    & 2.089    & 2.078 & 16.679   \\
    Queries     & 150528    & 137    & 256 & 457    & 900 \\
   
    \hline
    \end{tabular}
    }
    \caption{The effectiveness of SPGQ and finite difference query methods. 
    % Variance is a metric to measure the impact of architectures. 
    % Our method with smallest variance means using different architectures does not make a big difference.
    }
    \label{tab:sg}
\end{table*}

\section{More study about SGP}
Empirical validation of SGP's efficacy is demonstrated through two categories of experiments. On one hand, knowledge distillation experiments were conducted. When the student model distills unpurified sample gradients and logits knowledge, the resultant accuracy exhibits a decline due to irregular variance in the sample gradients, as compared to training without distillation. Conversely, distillation using purified sample gradients in conjunction with logits culminates in accuracy surpassing that achieved by distilling logits alone. On the other hand, T-SNE visualization was employed, concatenating the model-extracted sample features with the purified sample gradients. Comparative analysis reveals that purified sample gradients significantly enhance the final visualization outcome, as opposed to scenarios involving no concatenation or concatenation with unpurified sample gradients. The aforementioned experiments collectively attest to the effectiveness of the purification mechanism in eliminating variance from sample gradients.

\subsection{knowledge distillation}
\begin{figure*}[h]
    \centering
    \includegraphics[width=0.9\textwidth]{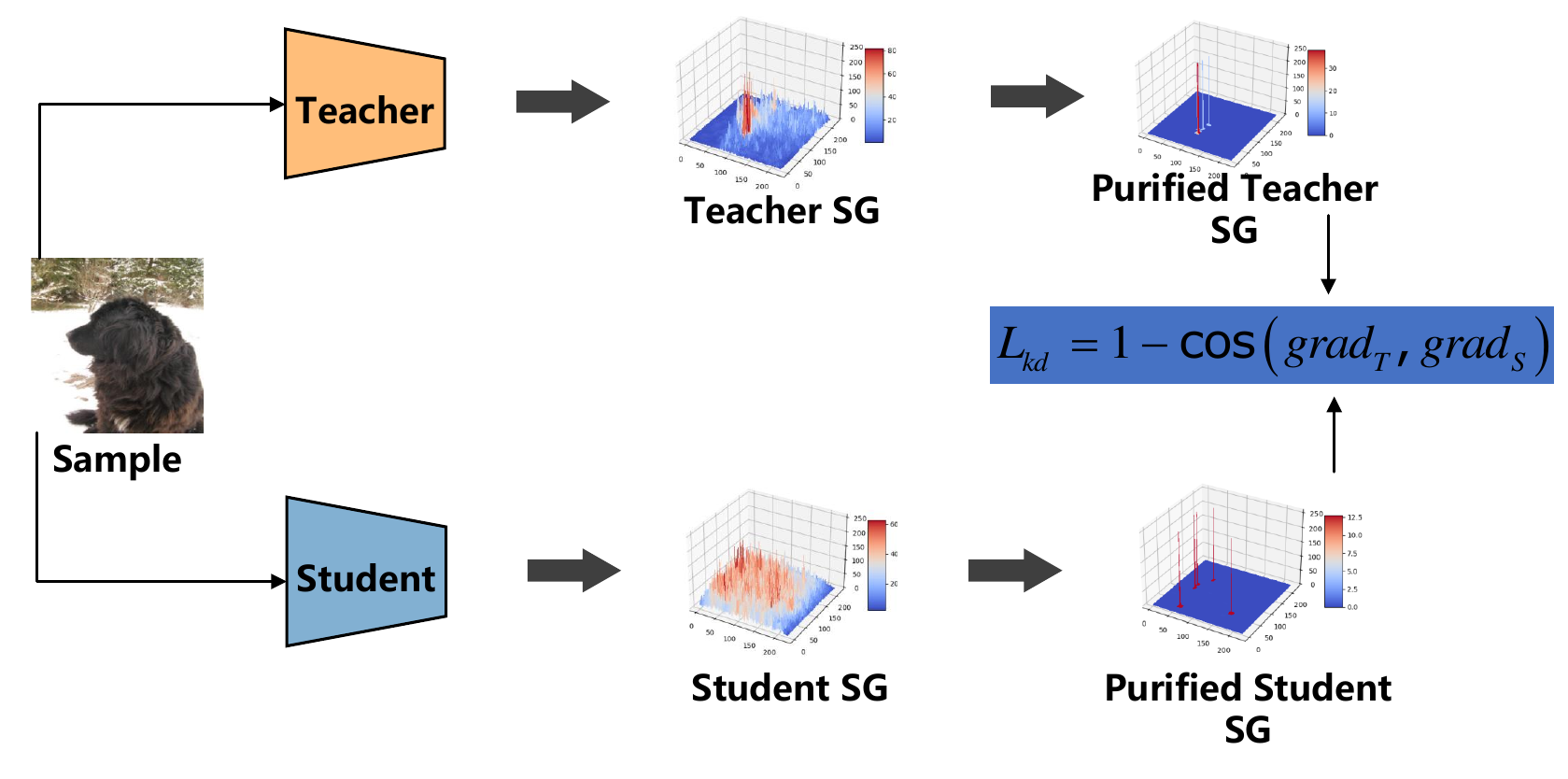}
    \caption{The framework of sample gradient knowledge distillation with SGP}
    \label{kdd}
\end{figure*}

We conducted experiments on image classification knowledge distillation. As shown in Figure \ref{kdd}, the sample gradients obtained from passing the samples through the teacher and student models are processed by SGP and then associated through the loss function for distillation. We selected ICKD \cite{ICKD}, Overhaul \cite{overhaul}, AT \cite{AT}, FitNet \cite{Fitnet}, and FSP \cite{FSP}, KD \cite{kd}, RKD \cite{RKD}, DIST \cite{DIST}, SRRL \cite{SRRL}, and CRD \cite{CRD} as the baselines. Initially, we compared the effects of each baseline with or without SGKD used individually in CIFAR100 \cite{cifar100}. The training strategies for CIFAR100 is shown in Table \ref{Strategies}. We selected a series of teacher-student model combinations from VGG \cite{vgg}, ResNet \cite{resnet}, and their variants \cite{wrn}. CIFAR100 Experimental Results: As depicted in Table \ref{cifar}, using SGP individually results in improved accuracy for student models. To further examine the influence of SGP on the knowledge distillation task, we compare the accuracy of the student model with and without SGP on the CIFAR100 dataset. As demonstrated in Table \ref{ablation}, the considerable numerical difference between the original sample gradients of the student and teacher hinders the accurate transfer of the teacher model's dark knowledge to the student model. This results in a decrease in the student model's performance. With SGP, the student model can effectively learn the teacher's dark knowledge.

\subsection{T-SNE visualization}

\begin{figure*}[htbp]
    \centering
    \begin{subfigure}[b]{0.24\textwidth}
    \includegraphics[width=\textwidth]{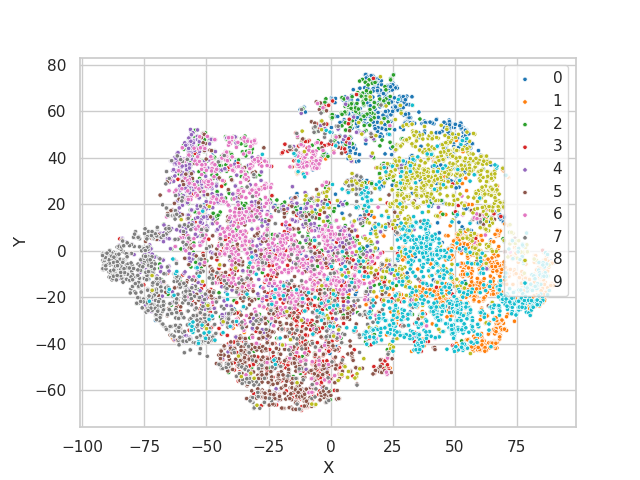}
    \end{subfigure}
    \begin{subfigure}[b]{0.24\textwidth}
    \includegraphics[width=\textwidth]{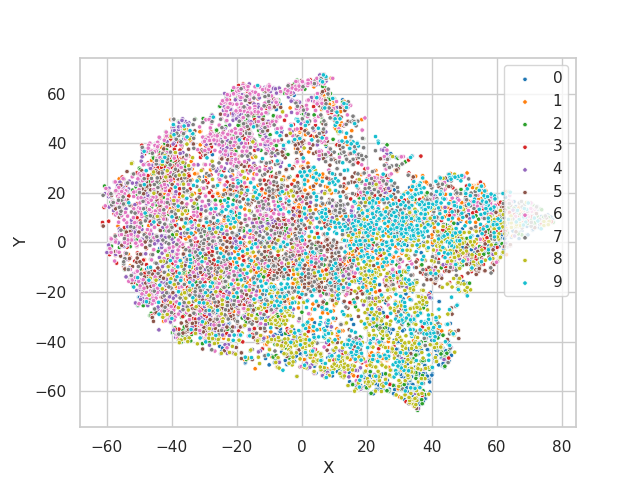}
    \end{subfigure}
    \begin{subfigure}[b]{0.24\textwidth}
    \includegraphics[width=\textwidth]{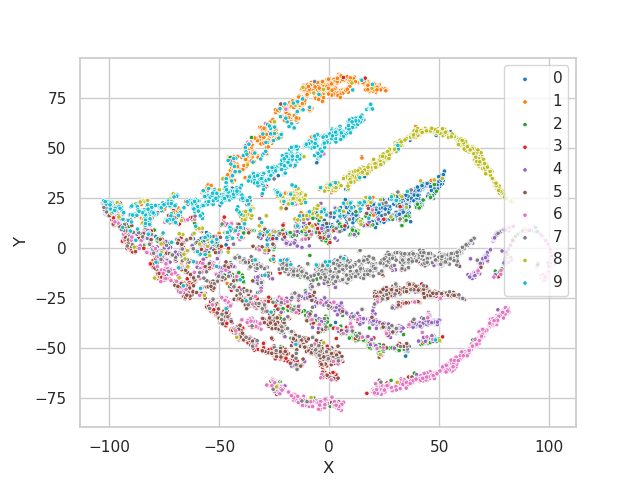}
    \end{subfigure}
    \begin{subfigure}[b]{0.24\textwidth}
    \includegraphics[width=\textwidth]{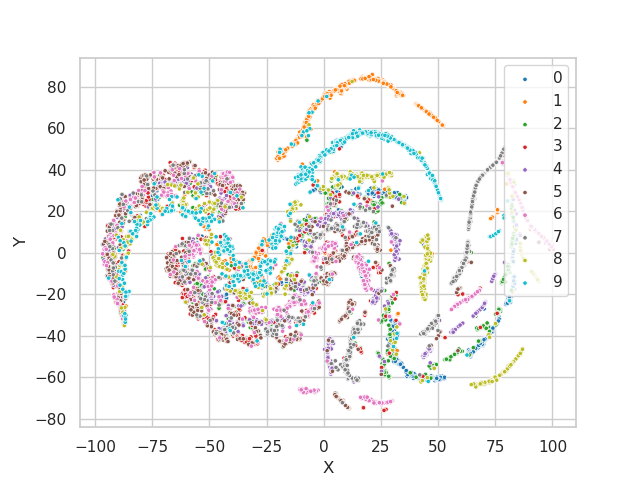}
    \end{subfigure}
    \caption{The images from left to right are original feature vector classification, feature vector classification with original SG, feature vector classification with purified SG as \(\beta = 0.5\), and classification with purified SG as \(\beta = 0.2\).}
    \label{fig:ablation}
\end{figure*}

SGP allows pixel-level sample gradients to possess more class information. We first train a resnet34 on CIFAR-10. Then, we obtain the test sample feature vectors through the final layer before the output of resnet34. The feature vectors are visualized using T-SNE. Next, we concatenate the purified sample gradients or the original sample gradients behind the feature vectors and visualize them again. The four visualization results show that the sample gradients purified by SGP can effectively aid in classification. In contrast, the original sample gradients, containing variance and having low informational content, provide no benefit to the representation of feature vectors. The result is shown in Figure \ref{fig:ablation}.

\begin{table*}[htbp]
\centering
\caption{
The agreement (in \%), test accuracy (in \%), and queries of each method with hard label. For our model, we report the average result as well as the standard deviation computed over 5 runs. (\textbf{Boldface}: the best value.)}
\resizebox{\textwidth}{!}{
\begin{tabular}{lccccccccc}
\hline
\multirow{2}{*}{Method (hard-label)}    & \multicolumn{3}{c}{Indoor (10k)}         & \multicolumn{3}{c}{Indoor (15k)}            & \multicolumn{3}{c}{Indoor (20k)}\\
\cline{2-10}
                           & Agreement        & Acc    &Queries           & Agreement        & Acc      &Queries        & Agreement        & Acc       &Queries       \\
\hline
ZSDB3KD       & 27.55          & 26.43          & 1109k          & 29.52         & 30.07          & 1002k  & 34.21          & 33.71   &1229k\\
DFMS      & 28.75          & 27.13          & 1321k          & 30.12          & 29.35          & 993k  & 34.23          & 33.15    &989k\\
EDFBA          & 27.56          & 26.55          & 345k          & 30.34          & 29.48          & 477k  & 34.12          & 33.72     &531k\\
DS & 27.52          & 26.55          & 1200k          & 30.53          & 29.88          & 1090k  & 35.24          & 34.18     &996k\\
knockoff                     & 25.31 & 23.66 & 10k & 27.19 & 25.73 & 15k  &  31.23    & 29.93    &20k\\
ActiveThief                     & 25.01 & 24.19 & 10k & 27.59 & 26.13 & 15k  &  30.98    & 30.11    &20k\\
Black-Box Dissector                     & 25.91 & 23.57 & 20k & 27.43 & 26.26 & 30k  &  31.59    &30.46    &40k\\
SPSG(Ours)              & \textbf{27.86}$\pm$0.16 & \textbf{26.79}$\pm$0.21 & 132k$\pm$0.01k & \textbf{31.43.27}$\pm$0.34 & \textbf{30.29}$\pm$0.13 & 195k$\pm$0.01k &\textbf{38.27}$\pm$0.34 & \textbf{36.32}$\pm$0.13  &371k$\pm$0.01k\\
\hline
\end{tabular}
}
\label{H_dataresult}
\end{table*}
\begin{table*}[htbp]
\centering
\caption{
The agreement (in \%) and test accuracy (in \%)s of each method with different proxy architecture \cite{resnet,vgg,Densenet} in CUBS-200-2011. The real sample number is 20k. For our model, we report the average result as well as the standard deviation computed over 5 runs. (\textbf{Boldface}: the best value.)}
\resizebox{\textwidth}{!}{
\begin{tabular}{lcccccccc}
\hline
\multirow{2}{*}{Method (probability)}    & \multicolumn{2}{c}{ResNet-18}         & \multicolumn{2}{c}{ResNet-50}            & \multicolumn{2}{c}{VGG-16} & \multicolumn{2}{c}{DenseNet}\\
\cline{2-9}
                           & Agreement        & Acc     & Agreement        & Acc         & Agreement        & Acc        & Agreement        & Acc       \\
\hline
ZSDB3KD       & 48.55          & 47.23               & 51.51         & 50.04     & 48.11          & 47.61 &  51.23    & 50.93  \\
DFMS      & 49.55          & 48.11          & 51.46         & 50.25      & 48.63          & 47.85 &  52.13    & 51.93 \\
EDFBA          & 49.76          & 48.55          & 51.32          & 50.11     & 48.72          & 47.84 &  51.13    & 49.98\\
DS & 48.72          & 46.59              & 50.77          & 50.23         & 48.95          & 48.22     &  50.33    & 49.71 \\
knockoff                     & 45.39 & 43.98  & 47.28 & 45.63  &  49.78    & 48.23 &  50.21    & 49.28\\
ActiveThief                     & 47.01 & 46.19  & 48.59 & 46.19   &  50.18    & 50.11 &  50.23    & 49.13\\
Black-Box Dissector                     & 46.77 & 45.87  & 48.49 & 47.56   &  50.19    &50.16  &  51.23    & 50.93\\
SPSG(Ours)              & \textbf{49.96}$\pm$0.16 & \textbf{49.70}$\pm$0.19  & \textbf{52.27}$\pm$0.34 & \textbf{51.39}$\pm$0.19  &\textbf{51.21}$\pm$0.14 & \textbf{51.12}$\pm$0.11  &\textbf{52.71}$\pm$0.34 & \textbf{52.32}$\pm$0.13\\
\hline
\end{tabular}
}
\label{a_dataresult}
\end{table*}

\begin{table*}[htbp]
\centering
\caption{
The agreement (in \%) and test accuracy (in \%) of each method with different sample selection Strategies in CUBS-200-2011. For our model, we report the average result as well as the standard deviation computed over 5 runs. (\textbf{Boldface}: the best value.)}
\resizebox{\textwidth}{!}{
\begin{tabular}{lcccccccc}
\hline
\multirow{2}{*}{Method (probability)}    & \multicolumn{2}{c}{CUBS200(10k)}         & \multicolumn{2}{c}{CUBS200(15k)}            & \multicolumn{2}{c}{CUBS200(20k)} & \multicolumn{2}{c}{CUBS200(25k)}\\
\cline{2-9}
                           & Agreement        & Acc     & Agreement        & Acc         & Agreement        & Acc        & Agreement        & Acc       \\
\hline
knockoff (K-center)       & 55.65          & 52.71               & 59.32         & 55.71     & 61.77          & 57.61 &  63.21    & 61.93  \\
knockoff (Reinforce)       & 54.37          & 50.11          & 56.76         & 54.21      & 59.67          & 57.15 &  61.13    & 59.92 \\
Black-Box Dissector (K-center)   & 57.76          & 53.53          & 59.38          & 56.47     & 61.22          & 58.81 &  63.10    & 62.99\\
Black-Box Dissector (Reinforce) & 55.71          & 52.12             & 58.71          & 56.23         & 61.95          & 59.11     &  61.31    & 60.71 \\
SPSG (K-center)                     & 59.27$\pm$0.11 & \textbf{56.81}$\pm$0.11  &\textbf{61.49}$\pm$0.17 & \textbf{60.56}$\pm$0.14   &  63.17$\pm$0.14    &62.16$\pm$0.12    &\textbf{65.21}$\pm$0.31    & \textbf{63.93}$\pm$0.11\\
SPSG  (Reinforce)               & \textbf{59.96}$\pm$0.16 & 56.70$\pm$0.19  & 61.27 & 60.39$\pm$0.19  &\textbf{63.23}$\pm$0.14 & \textbf{62.19}$\pm$0.11  &64.11$\pm$0.34 & 62.32$\pm$0.13\\
\hline
\end{tabular}
}
\label{sta_dataresult}
\end{table*}

\begin{table*}[h]
  \centering
  \caption{Strategies for CIFAR100}
  \resizebox{1\linewidth}{!}{
    \begin{tabular}{c|cccccccc}
    Strategy & Dataset & Epochs & Batch size & Initial LR & Optimizer & Weight decay & LR scheduler & Data augmentation \\
    \midrule
    A1    & CIFAR-100 & 240   & 64    & 0.05  & SGD   & 0.0005 & X0.1 at 150,180,210 epochs & crop+flip \\
    \bottomrule
    \end{tabular}}%
  \label{Strategies}%
\end{table*}%

\begin{table*}[h]
  \centering
  \caption{Results (accuracy:$\%$) in CIFAR100}
   \resizebox{0.8\textwidth}{!}{
    \begin{tabular}{c|cccccc}
    \toprule
    \multicolumn{1}{p{4.19em}|}{Teacher\newline{}Student} & \multicolumn{1}{p{5em}}{WRN-40-2\newline{}WRN-16-2} & \multicolumn{1}{p{5.94em}}{WRN-40-2\newline{}WRN-16-2\newline{}(with SGP)} & \multicolumn{1}{p{4.69em}}{ResNet110\newline{}ResNet20} & \multicolumn{1}{p{6em}}{ResNet110\newline{}ResNet20\newline{}(with SGP)} & \multicolumn{1}{p{3.75em}}{Vgg13\newline{}Vgg8} & \multicolumn{1}{p{6.5em}}{Vgg13\newline{}Vgg8\newline{}(with SGP)} \\
    \midrule
    \multicolumn{1}{p{4.19em}|}{Teacher} & 75.61 & /     & 74.31 & /     & 74.64 & / \\
    \multicolumn{1}{p{4.19em}|}{Student} & 73.26 & /     & 69.06 & /     & 70.36 & / \\
    \midrule
    kD \cite{kd} & $74.92\pm1.1$ & $75.48\pm1.3$\textcolor{green}{$\uparrow$} & $70.67\pm2.5$ & $71.78\pm1.2$\textcolor{green}{$\uparrow$} & $72.98\pm1.2$ & $72.78\pm1.1$\textcolor{red}{$\downarrow$} \\
    RKD \cite{RKD} & $73.35\pm2.2$ & $74.97\pm2.1$\textcolor{green}{$\uparrow$} & $69.25\pm1.6$ & $70.28\pm1.8$\textcolor{green}{$\uparrow$} & $71.48\pm1.4$ & $71.64\pm1.3$\textcolor{green}{$\uparrow$} \\
    DIST \cite{DIST} & $73.78\pm1.3$ & $74.79\pm1.4$\textcolor{green}{$\uparrow$} & $71.86\pm1.2$ & $71.91\pm3.5$\textcolor{green}{$\uparrow$} & $71.62\pm1.7$ & $71.67\pm1.6$\textcolor{green}{$\uparrow$}\\
    SRRL \cite{SRRL} & $73.71\pm1.7$ & $75.16\pm3.1$\textcolor{green}{$\uparrow$} & $70.91\pm1.4$ & $71.23\pm2.7$\textcolor{green}{$\uparrow$} & $71.45\pm1.8$ & $71.78\pm2.1$\textcolor{green}{$\uparrow$} \\
    CRD \cite{CRD} & $75.48\pm2.1$ & $75.61\pm1.1$\textcolor{green}{$\uparrow$} & $71.46\pm1.7$ & $71.57\pm3.6$\textcolor{green}{$\uparrow$} & $73.94\pm1.2$ & $73.62\pm2.3$\textcolor{red}{$\downarrow$} \\
    SGKD   & $74.91\pm1.1$ & /     & $71.48\pm1.7$ & / & $72.61\pm2.1$ & / \\
    ICKD \cite{ICKD} & $75.34\pm1.2$ & $75.44\pm1.6$\textcolor{green}{$\uparrow$} & $71.91\pm1.3$ & $72.01\pm2.1$\textcolor{green}{$\uparrow$} & $73.88\pm2.2$ & $74.09\pm1.3$\textcolor{green}{$\uparrow$} \\
    overhaul \cite{overhaul}& $75.52\pm1.3$ & $75.48\pm1.4$\textcolor{red}{$\downarrow$} & $71.21\pm1.5$ & $71.34\pm1.2$\textcolor{green}{$\uparrow$} & $73.42\pm1.1$ & $73.57\pm1.9$\textcolor{green}{$\uparrow$} \\
    AT  \cite{AT}  & $74.08\pm1.7$ & $74.61\pm2.3$\textcolor{green}{$\uparrow$} & $70.22\pm2.1$ & $71.24\pm1.8$\textcolor{green}{$\uparrow$} & $71.43\pm1.9$ & $72.86\pm2.1$\textcolor{green}{$\uparrow$} \\
    FitNet \cite{Fitnet} & $73.58\pm2.3$ & $73.67\pm2.1$\textcolor{green}{$\uparrow$} & $68.99\pm1.2$ & $71.67\pm2.3$\textcolor{green}{$\uparrow$} & $71.02\pm2.5$ & $72.21\pm2.4$\textcolor{green}{$\uparrow$} \\
    FSP \cite{FSP}  & $72.91\pm2.1$ & $73.28\pm1.7$\textcolor{green}{$\uparrow$} & $70.11\pm1.5$ & $71.94\pm1.5$\textcolor{green}{$\uparrow$} & $70.23\pm1.3$ & $71.47\pm1.5$\textcolor{green}{$\uparrow$} \\
    \bottomrule
    \end{tabular}}%
  \label{cifar}%
\end{table*}%

\begin{table*}[htbp]
  \centering
  \caption{Ablation study on CIFAR100.}
   \resizebox{0.8\textwidth}{!}{
    \begin{tabular}{c|cccccc}
    \toprule
    \multicolumn{1}{p{4.19em}|}{Teacher\newline{}Student} & \multicolumn{1}{p{5em}}{WRN-40-2\newline{}WRN-16-2\newline{}(with SG)} & \multicolumn{1}{p{5.94em}}{WRN-40-2\newline{}WRN-16-2\newline{}(with SGP)} & \multicolumn{1}{p{4.69em}}{ResNet110\newline{}ResNet20\newline{}(with SG)} & \multicolumn{1}{p{6em}}{ResNet110\newline{}ResNet20\newline{}(with SGP)} & \multicolumn{1}{p{6.5em}}{Vgg13\newline{}Vgg8\newline{}(with SG)} & \multicolumn{1}{p{6.5em}}{Vgg13\newline{}Vgg8\newline{}(with SGP)} \\
    \midrule
    \multicolumn{1}{p{4.19em}|}{Teacher} & 75.61 & /     & 74.31 & /     & 74.64 & / \\
    \multicolumn{1}{p{4.19em}|}{Student} & 73.26 & /     & 69.06 & /     & 70.36 & / \\
    \midrule
    KD\cite{kd} & $72.13\pm1.4$\textcolor{red}{$\downarrow$} & $75.48\pm1.3$\textcolor{green}{$\uparrow$} & $68.77\pm2.1$\textcolor{red}{$\downarrow$} & $71.78\pm1.2$\textcolor{green}{$\uparrow$} & $69.58\pm1.2$\textcolor{red}{$\downarrow$} & $72.78\pm1.1$\textcolor{red}{$\downarrow$} \\
    RKD \cite{RKD}& $71.39\pm2.0$\textcolor{red}{$\downarrow$} & $74.97\pm2.1$\textcolor{green}{$\uparrow$} & $68.45\pm2.6$\textcolor{red}{$\downarrow$} & $70.28\pm1.8$\textcolor{green}{$\uparrow$} & $69.18\pm1.2$\textcolor{red}{$\downarrow$} & $71.64\pm1.3$\textcolor{green}{$\uparrow$} \\
    DIST \cite{DIST}& $72.44\pm1.2$\textcolor{red}{$\downarrow$} & $74.79\pm1.4$\textcolor{green}{$\uparrow$} & $68.76\pm1.3$\textcolor{red}{$\downarrow$} & $71.91\pm3.5$\textcolor{green}{$\uparrow$} & $69.52\pm1.3$\textcolor{red}{$\downarrow$} & $71.67\pm1.6$\textcolor{green}{$\uparrow$}\\
    SRRL \cite{SRRL}& $73.16\pm1.2$\textcolor{red}{$\downarrow$} & $75.16\pm3.1$\textcolor{green}{$\uparrow$} & $68.45\pm1.8$\textcolor{red}{$\downarrow$} & $71.23\pm2.7$\textcolor{green}{$\uparrow$} & $70.16\pm1.4$\textcolor{red}{$\downarrow$} & $71.78\pm2.1$\textcolor{green}{$\uparrow$} \\
    CRD \cite{CRD}& $72.18\pm1.8$\textcolor{red}{$\downarrow$} & $75.61\pm1.1$\textcolor{green}{$\uparrow$} & $68.66\pm2.7$\textcolor{red}{$\downarrow$} & $71.57\pm3.6$\textcolor{green}{$\uparrow$} & $70.11\pm1.2$\textcolor{red}{$\downarrow$} & $73.62\pm2.3$\textcolor{red}{$\downarrow$} \\
    ICKD \cite{ICKD} & $72.39\pm1.2$\textcolor{red}{$\downarrow$} & $75.44\pm1.6$\textcolor{green}{$\uparrow$} & $71.91\pm1.5$\textcolor{red}{$\downarrow$} & $72.01\pm2.1$\textcolor{green}{$\uparrow$} & $68.34\pm2.7$\textcolor{red}{$\downarrow$} & $74.09\pm1.3$\textcolor{green}{$\uparrow$} \\
    overhaul \cite{overhaul}& $71.98\pm2.1$\textcolor{red}{$\downarrow$} & $75.48\pm1.4$\textcolor{red}{$\downarrow$} & $71.21\pm1.3$\textcolor{red}{$\downarrow$} & $71.34\pm1.2$\textcolor{green}{$\uparrow$} & $69.67\pm1.4$\textcolor{red}{$\downarrow$} & $73.57\pm1.9$\textcolor{green}{$\uparrow$} \\
    AT \cite{AT}   & $72.78\pm1.1$\textcolor{red}{$\downarrow$} & $74.61\pm2.3$\textcolor{green}{$\uparrow$} & $70.22\pm2.7$\textcolor{red}{$\downarrow$} & $71.24\pm1.8$\textcolor{green}{$\uparrow$} & $69.89\pm1.5$\textcolor{red}{$\downarrow$} & $72.86\pm2.1$\textcolor{green}{$\uparrow$} \\
    FitNet\cite{Fitnet} & $72.18\pm1.2$\textcolor{red}{$\downarrow$} & $73.67\pm2.1$\textcolor{green}{$\uparrow$} & $68.99\pm1.2$\textcolor{red}{$\downarrow$} & $71.67\pm2.3$\textcolor{green}{$\uparrow$} & $69.78\pm2.1$\textcolor{red}{$\downarrow$} & $72.21\pm2.4$\textcolor{green}{$\uparrow$} \\
    FSP \cite{FSP}  & $72.11\pm2.1$\textcolor{red}{$\downarrow$} & $73.28\pm1.7$\textcolor{green}{$\uparrow$} & $70.11\pm1.3$\textcolor{red}{$\downarrow$} & $71.94\pm1.5$\textcolor{green}{$\uparrow$} & $69.35\pm1.3$\textcolor{red}{$\downarrow$} & $71.47\pm1.5$\textcolor{green}{$\uparrow$} \\
    \bottomrule
    \end{tabular}}%
  \label{ablation}%
\end{table*}

